\documentclass[10pt,twocolumn,letterpaper]{article}

\usepackage{cvpr}              % To produce the CAMERA-READY version
\definecolor{cvprblue}{rgb}{0.21,0.49,0.74}
\usepackage[pagebackref,breaklinks,colorlinks,citecolor=cvprblue, linkcolor=red]{hyperref}
\usepackage{multirow}
\usepackage[dvipsnames]{xcolor}
\usepackage{ulem} 
\usepackage{pifont}
\usepackage{float}
\usepackage{indentfirst}
\usepackage{amsmath}
\usepackage{makecell}%To keep spacing of text in tables
\setcellgapes{4pt}
\usepackage{ulem}
\usepackage{makecell}
\usepackage{utfsym}
\usepackage[accsupp]{axessibility}
\usepackage{lineno}
\def\paperID{9676} % *** Enter the Paper ID here
\def\confName{CVPR}
\def\confYear{2025}

\title{UniPose: A Unified Multimodal Framework for Human \\ Pose Comprehension, Generation and Editing}

\author{%
  \small \textbf{Yiheng Li}$^{1,2}$, \textbf{Ruibing Hou}$^{1}$\thanks{Corresponding author} \;, \textbf{Hong Chang}$^{1,2}$, \small \textbf{Shiguang Shan}$^{1,2}$ \textbf{,} \textbf{Xilin Chen}$^{1,2}$\\ 
  \small {$^1$Key Laboratory of Intelligent Information Processing of Chinese Academy of Sciences (CAS),} \\ \small {Institute of Computing Technology, CAS, China} \\
  \small {$^2$University of Chinese Academy of Sciences, China} \\
  \small \texttt{yiheng.li@vipl.ict.ac.cn,\{houruibing,changhong,sgshan,xlchen\}@ict.ac.cn}\\
}

\begin{document}

\twocolumn[{
\renewcommand\twocolumn[1][]{#1}
\maketitle
\vspace{-3em}
\begin{center}
    \captionsetup{type=figure}
    \includegraphics[width=0.95\textwidth]{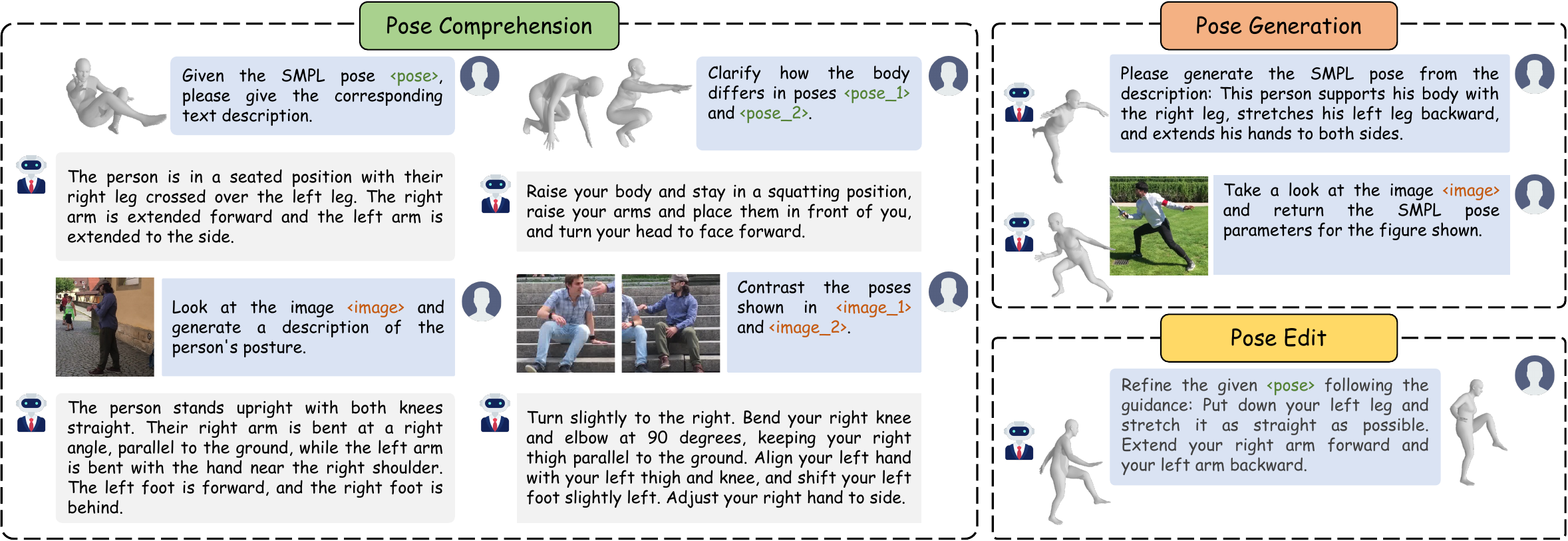}
    \captionof{figure}{UniPose can handle pose comprehension, generation and editing tasks under different instructions within a unified framework.}
\end{center}
}]

\begin{abstract}
\vspace{-2em}

Human pose plays a crucial role in the digital age. While recent works have achieved impressive progress in understanding and generating human poses, they often support only a single modality of control signals and operate in isolation, limiting their application in real-world scenarios. 
This paper presents UniPose, a framework employing Large Language Models (LLMs) to comprehend, generate, and edit human poses across various modalities, including images, text, and 3D SMPL poses. Specifically,  we apply a pose tokenizer to convert 3D poses into discrete pose tokens, enabling seamless integration into the LLM within a unified vocabulary. To further enhance the fine-grained pose perception capabilities, 
we facilitate UniPose with a mixture of visual encoders, among them a pose-specific visual encoder.
Benefiting from a unified learning strategy, UniPose effectively transfers knowledge across different pose-relevant tasks, adapts to unseen tasks, and exhibits extended capabilities.
This work serves as the first attempt at building a general-purpose framework for pose comprehension, generation, and editing. 
Extensive experiments highlight UniPose's competitive and even superior performance across various pose-relevant tasks. 
Code is available at \href{https://github.com/liyiheng23/UniPose}{https://github.com/liyiheng23/UniPose}.
\vspace{-2em}
\end{abstract}

\section{Introduction}\label{sec:intro}
\vspace{-2mm}

\begin{table*}[]
\resizebox{\textwidth}{!}{
\begin{tabular}{l|cccc|cc|c}
  \toprule
  \multirow{2}{*}{Tasks}& \multicolumn{4}{c|}{Pose Comprehension} & \multicolumn{2}{c|}{Pose Generation} & \multirow{2}{*}{Pose Editing}\\ \cline{2-7} 
   & Pose-to-Text & Image-to-Text & Pose-Diff & Image-Diff & Text-to-Pose & Pose Estimation & \\
  \midrule
  Input$\rightarrow$Output & Pose$\rightarrow$Text & Image$\rightarrow$Text & Pose Pair$\rightarrow$Text & Image Pair$\rightarrow$Text & Text$\rightarrow$Pose & Image$\rightarrow$Pose & Pose\&Text$\rightarrow$Pose \\
  \midrule
  HMR 2.0 \cite{goel2023hmr2.0} & \ding{56} & \ding{56} & \ding{56} & \ding{56} & \ding{56} & \ding{52} & \ding{56} \\
  PoseScript \cite{delmas2024posescript} & \ding{52} & \ding{56} & \ding{56} & \ding{56} & \ding{52} & \ding{56} & \ding{56} \\
  PoseFix \cite{delmas2023posefix} & \ding{56} & \ding{56} & \ding{52} & \ding{56} & \ding{56} & \ding{56} & \ding{52} \\
  ChatPose \cite{feng2024chatpose} & \ding{56} & \ding{52} & \ding{56} & \ding{56} & \ding{52} & \ding{52} & \ding{56} \\
  ChatHuman \cite{lin2024chathuman} & \ding{56} & \ding{52} & \ding{56} & \ding{56} & \ding{52} & \ding{52} & \ding{56} \\
  PoseEmbroider \cite{delmas2024poseembroider} & \ding{52} & \ding{52} & \ding{52} & \ding{52} & \ding{52} & \ding{52} & \ding{56} \\
  \midrule
  UniPose (Ours) & \ding{52} & \ding{52} & \ding{52} & \ding{52} & \ding{52} & \ding{52} & \ding{52} \\
  \bottomrule
\end{tabular}
}
\caption{Comparison of recent methods across various pose comprehension, generation and editing tasks. }
\label{tab-pose-tasks}
\vspace{-7mm}
\end{table*}

Human pose plays a pivotal role in various human-centric applications such as VR and healthcare. 
Numerous studies focus on \textit{single-pose comprehension}, \textit{i.e.}, producing posture-relevant description from a 3D body pose \cite{delmas2024posescript} or human image \cite{feng2024chatpose}, as well as \textit{pose generation}, \textit{i.e.}, creating complex 3D poses from textual descriptions \cite{delmas2024posescript, lin2023being, hong2022avatarclip} or human images \cite{dwivedi2024tokenhmr, sun2024aios, xu2023smpler, cai2024smplerx, fiche2024vq-hps}. 
Recently, a few studies have explored the relationship between pairs of poses \cite{delmas2023posefix, fieraru2021aifit, kim2021fixmypose}. These studies investigate \textit{pose-pair comprehension}, where textual instruction is produced based on the differences between two 3D poses, and \textit{pose editing}, where corrected 3D body pose is generated based on an initial pose and modification instruction. 
However, a key limitation of existing work is that pose comprehension, generation, and editing are predominantly studied in isolation. In reality, human pose cognition and communication inherently involve seamless transitions between multiple pose-relevant modalities, including 3D SMPL poses \cite{smpl}, textual descriptions, and human images. This highlights the need for a unified multimodal framework capable of simultaneously handling pose comprehension, generation, and editing.
% Therefore, it is essential to develop a unified multimodal framework capable of simultaneously addressing pose comprehension, generation, and editing.

Recent years have witnessed a significant breakthrough in large language models (LLMs) \cite{wei2024occllama, gunasekar2023phi, jiang2023mistral} and multimodal LLMs (MLLMs), enabling general-purpose analysis of images \cite{liu2023llava, bai2023qwen-vl}, videos \cite{lin2023video-llava, ye2024mPLUG-Owl3}, motions \cite{jiang2023motiongpt, chen2024motionllm-video, zhang2024motiongptaaai}, and audios \cite{yang2024uniaudio, zhang2023speechgpt}. 
In the area of human poses, ChatPose \cite{feng2024chatpose}, a recent innovation, leverages LLMs to generate 3D human poses from images and textual descriptions. Nevertheless, it focuses solely on single-pose generation, lacking the capacity for pose comprehension and editing.  
Moreover, existing MLLMs still fall short in providing comprehensive analysis of human poses, particularly concerning fine-grained part semantics and complex relationships between pose pairs. 
Consequently, a unified multimodal LLM that enables finer-grained pose comprehension, generation, and complex pose editing is still in highly demand.
 % Moreover, ChatPose adopts non-unified processing of 3D poses and texts, encoding 3D poses as continuous high-level features while tokenizing linguistic texts into discrete token sequences. This non-unified processing introducing an extra burden for LLMs to model the interaction between 3D poses and texts, thus complicates the unification of pose comprehension, generation and editing.

Two main challenges need to be solved for building such a unified multimodal LLM framework.
The first challenge is creating a unified representation space across 3D poses and texts, enabling the unification of diverse pose-relevant tasks.
Existing work \cite{feng2024chatpose} processes 3D poses and texts differently, encoding 3D poses as continuous high-level features while tokenizing linguistic texts into discrete token sequences. This non-unified processing incurs an extra burden on LLMs to model interactions between 3D poses and texts, hindering the unifying of pose comprehension, generation and editing.
The second challenge lies in achieving fine-grained pose perception within the visual branch of the multimodal framework. Most MLLMs \cite{liu2023llava, bai2023qwen-vl, wang2023cogvlm, feng2024chatpose} employ CLIP \cite{radford2021clip} as their visual branch. While CLIP's visual encoder aligns well with the text embedding space through image-text contrastive learning, it struggles to capture detailed pixel-level information, such as keypoints and parsing maps, due to the global supervision provided by image captions. This limitation constrains MLLM's capabilities in fine-grained pose comprehension and generation.

To address these challenges, we propose UniPose, a uniform multimodal framework for human pose comprehension, generation and editing, which harnesses the powerful language generation abilities of LLMs to unify various pose-relevant tasks (Tab. \ref{tab-pose-tasks}). UniPose comprises three tires. 
\textbf{Firstly}, UniPose is equipped with a \textit{pose tokenizer} for processing 3D poses and texts uniformly.
Inspired by the observation that human poses exhibit a semantic coupling similar to language \cite{jiang2023motiongpt, luo2024m3GPT, wu2024motionllm},
%we treat 3D pose as a specific language. %to seamlessly integrate with text through LLM. 
%Specifically, akin to language, the pose tokenizer compresses raw 3D pose into a sequence of discrete semantic tokens.
we treat 3D pose as a specific language. Akin to language, the pose tokenizer compresses raw 3D pose into a sequence of discrete semantic tokens.
By encoding both 3D pose and language within a shared vocabulary, 
we build a unified representation space across 3D poses and texts, 
which enables LLMs to be easily adapted to handle pose comprehension, generation, and editing. 
% \uline{However, a unified vocabulary alone is insufficient to bridge the underlying gap between pose and text modalities. Thus, \textbf{secondly}, to address the distinct internal logical relationships between pose tokens and text tokens, we implement a mixed-attention mechanism within LLM.} 
\textbf{Secondly}, unlike most MLLMs \cite{liu2023llava, bai2023qwen-vl, wang2023cogvlm,feng2024chatpose} that solely rely on CLIP's visual encoder \cite{radford2021clip}, we adopt a mixture-of-visual-encoders that combines CLIP's original visual encoder with a pose-specific visual encoder pre-trained on pose estimation task.
This dual-encoder setup not only aligns visual representations with text embedding space but also enhances fine-grained pose perception, enabling more effective integration into the multimodal framework for improved pose comprehension and generation. 
\textbf{Thirdly}, we implement a mixed-attention mechanism within LLMs to handle the distinct internal logical relationships between pose and text tokens.
Unlike text tokens, pose tokens encode spatial joint positions without causal dependencies, making unified autoregressive modeling suboptimal.
To address this, we apply causal attention to text tokens and bidirectional attention to pose tokens. 
% This mixed-attention strategy not only preserves LLM's original reasoning capabilities but also enhances contextual pose perception, facilitating more effective pose manipulation and generation.
This mixed-attention strategy preserves LLM's original reasoning capabilities while enhancing contextual pose perception, enabling more effective pose generation and editing.
% Additionally, it enables parallel generation of all pose tokens in a single step, significantly accelerating inference.

To our knowledge, UniPose is the first approach to integrate seven core tasks of pose comprehension, generation, and editing into a uniform framework. 
Extensive experiments demonstrate that UniPose achieves competitive performance across multiple pose-relevant tasks. 
Additionally, through qualitative results, we demonstrate that UniPose possesses zero-shot generalization capabilities, \textit{e.g.}, text-enhanced pose estimation. 
\vspace{-3mm}

\section{Related Work}
\vspace{-1mm}
\noindent
\textbf{Human Pose Comprehension.} \
Pose comprehension involves generating natural language descriptions of human pose or differences between pose pairs. For single-pose comprehension, traditional methods classify basic human actions from images  \cite{zhao2017single_image_action_recog}, videos \cite{wang2024internvideo2, wang2023masked, wang2023videomae}, or skeletons data \cite{mondal2024hummuss, chen2021CTR-GC, foo2023unified-pose-model, ahn2023star-transformer}. However, these methods typically lack detailed descriptions of  specific body part positioning. 
To address this gap,  \cite{delmas2024posescript} introduces PoseScript dataset which pairs human poses with detailed body parts descriptions, and propose a pose-to-text generation model that uses cross-attention to embed pose information within a text transformer  for nuanced pose descriptions.
For pose-pair comprehension,  \cite{fieraru2021aifit, kim2021fixmypose, delmas2023posefix} describe differences between source and target poses based on images, videos, or 3D poses. 
For example, PoseFix \cite{delmas2023posefix} uses an MLP to fuse source and target pose, then uses cross-attention in a text transformer to generate descriptions of pose differences.
While these approaches enhance understanding of human poses from multimodal data, they are typically task-specific, with limited control conditions and application scenarios.

\noindent
\textbf{Human Pose Generation.} \
Pose generation synthesizes human poses conditioned on text or images. Text-conditioned pose generation  generally falls into  two categories: shape-oriented \cite{streuber2016bodytalk, gralnik2023semantify} and pose-oriented \cite{briq2021towards_better_adv_syn, hong2022avatarclip, lin2023being}, which generate 3D poses from descriptions of body attributes  (\textit{e.g.}, slim waist) and simple actions (\textit{e.g.}, running), respectively.
% For example,  AvatarCLIP \cite{hong2022avatarclip} retrieves poses by first constructing a pose codebook and then matching text features to similar entries in the codebook.
% given an action label, AvatarCLIP \cite{hong2022avatarclip} first constructs a pose codebook and then calculates the similarity between the text features and the the codebook entries, retrieving the corresponding pose.
Image-conditioned pose generation (also referred to pose estimation)  includes optimization-based and regression-based approaches. Optimization-based methods \cite{SMPL-X:2019, bogo2016keep_it_SMPL, fang2023learning_human_mesh_recovery, rempe2021humor, fan2021revitalizing_opt_for_3d_hps} iteratively estimate 3D pose parameters, ensuring the projection of predicted 3D joints aligns with 2D keypoints. Regression-based methods \cite{goel2023hmr2.0, dwivedi2024tokenhmr, kanazawa2018hmr, xu2023smpler, cai2024smplerx} %sun2024aios, pymaf2021, pymafx2023} 
use deep neural networks to directly predict 3D pose parameters from input images.
Although these methods have achieved promising results in pose generation, they  lack the capability of pose comprehension and editing.

\noindent
\textbf{Multimodal Large Language Models.}  \
Large Language Models (LLMs) \cite{touvron2023llama, jiang2023mistral, gunasekar2023phi, yang2024qwen2} have shown remarkable capabilities in textual comprehension and reasoning. %owing to their extensive datasets and large-scale architectures. 
These models have been adapted for multimodal tasks, leading to the development of multimodal large language models. 
% Researchers have leveraged the capabilities of LLMs to address multimodal tasks, expanding them to multimodal large language models (MLLMs). 
For example, models like mPLUG-Owl3 \cite{ye2024mPLUG-Owl3}, MiniGPT-4 \cite{zhu2023minigpt4} and LLaVA \cite{liu2023llava, liu2024improvedllava, li2024llava-onevision} uses a visual encoder to extract image features and a projection layer to align image embeddings with text  embeddings, enhancing general visual perception.
Moving towards task-specific applications, LISA \cite{lai2024lisa} and Video-LISA \cite{bai2024videolisa} extend MLLMs for segmentation by integrating SAM \cite{kirillov2023segmentanything} for generating fine-grained segmentation masks.
Additionally,  Show-o \cite{xie2024showo} and Transfusion \cite{zhou2024transfusion} combine MLLMs with diffusion models \cite{ho2020ddpm} to unify image understanding and generation. %Beyond vision, MLLMs have also expanded into 3D scene understanding \cite{xu2023pointllm, wei2024occllama, zhu2024llava3d} and audio processing \cite{yang2024uniaudio, zhang2023speechgpt}, further broadening their application scope. 
A recent work, ChatPose \cite{feng2024chatpose}, applies LLMs to pose-related tasks, aiming to build a versatile pose generator. However, it remains limited in its capacity for pose understanding and editing. 

\begin{figure*}[htbp]
    \centering
    \includegraphics[width=0.95\linewidth]{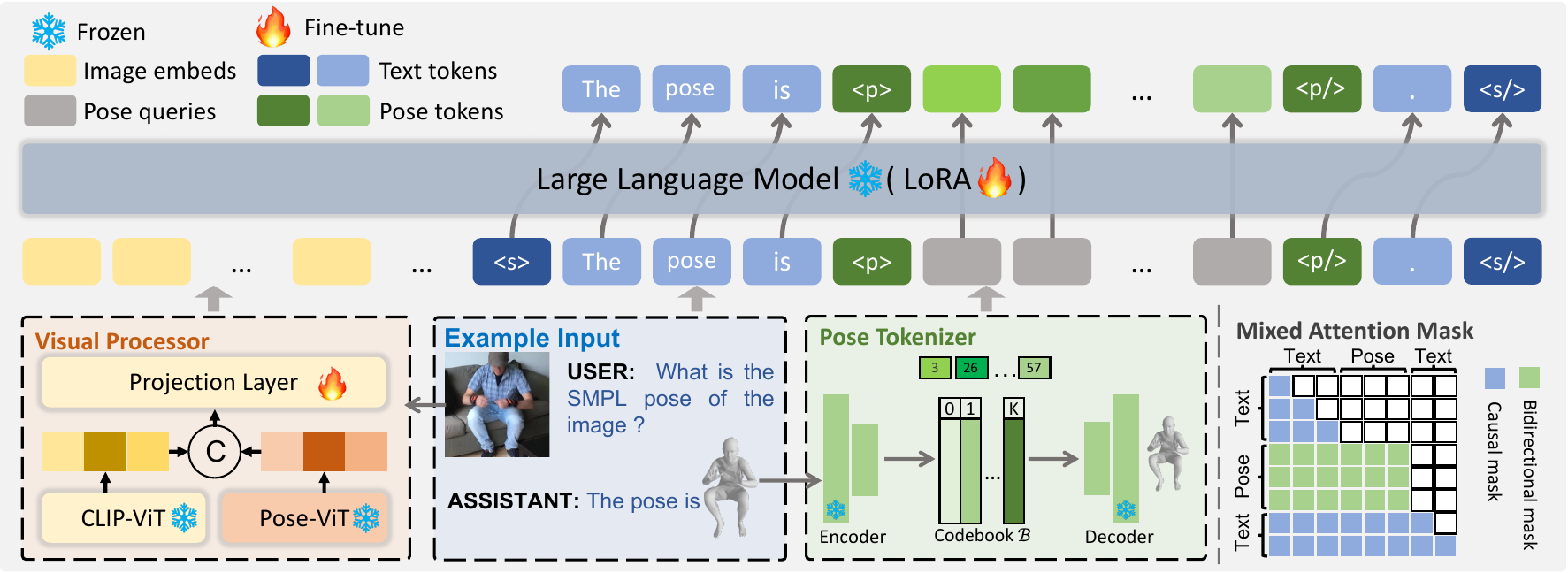}
    \caption{\textbf{Method overview}: UniPose comprises a Pose Tokenizer, Visual Processor and a pose-aware language LLM. Combining Pose Tokens learned by pose tokenizer, Visual Embeddings from visual processor and Text Tokens from text tokenizer, UniPose enables joint modeling of pose comprehension, generation and editing within a unified visual-language backbone.
    %The LLM backbone takes text tokens from the text tokenizer, pose tokens from the pose tokenizer, and image embeddings from the visual processor as input. The model then applies a mixed attention mechanism to jointly capture relationships across different modalities.
    }
    \label{fig:framework}
    % \vspace{-3mm}
    \vspace{-0.5cm}
\end{figure*} 
\vspace{-2mm}
\section{Method}
\vspace{-1mm}
To equip LLM with the capability to comprehend, generate, and edit human poses, we propose a unified framework named UniPose. As illustrated in Fig.~\ref{fig:framework}, UniPose comprises three main components: a pose tokenizer, which quantizes original 3D poses (represented as SMPL  \cite{smpl} pose parameters) into discrete tokens (Sec. \ref{pose-tokenizer}), a visual processor, which extracts fine-grained, pose-relevant features from visual inputs, and a pose-aware LLM, which supports unified modeling across multiple modalities (Sec. \ref{pose-vlm}).
To address pose-relevant tasks, we employ a four-stage straining scheme encompassing pose tokenizer training, pose-text alignment pre-training, vision projector pre-training, and instruction tuning (Sec. \ref{training-strategy}). During inference, pose tokens are decoded back to their original SMPL format by associated de-tokenizer, enabling various pose-relevant tasks to be executed via instructions (Sec. \ref{training-strategy})  

\vspace{-1mm}
\subsection{Pose Tokenizer}
\vspace{-1mm}
\label{pose-tokenizer}
% Previous work  \cite{feng2024chatpose} focuses solely on pose generation, where an LLM generates a single embedding  to represent a 3D pose, which is then mapped to SMPL parameters through multiple linear layers.  However, this single pose embedding cannot directly serves as input to LLM, restricting its ability to comprehend and edit poses. 
% To address this issue, we introduce a pose tokenizer that discretizes raw 3D pose into a sequence of language-like tokens, enabling the unification of pose and text within a single language model.  

To represent 3D pose in discrete tokens, we build the pose tokenizer based on Vector Quantized Variational Autoencoders (VQ-VAE) \cite{van2017vqvae}, as shown in Fig. \ref{fig:framework}. The pose tokenizer consists of an encoder $\mathcal{E}$, a decoder $\mathcal{D}$, along with a learnable codebook $\mathcal{B}_p = \{b_m\}_{m=1}^{M}$ containing $M$ discrete vectors. Formally, we represent a 3D pose $\boldsymbol{p}$ using SMPL pose parameters, \textit{i.e.}, $\boldsymbol{p}=\left[\boldsymbol{\gamma}, \boldsymbol{\theta}\right]$ where  $\boldsymbol{\gamma} \in \mathbb{R}^{6}$ denotes the root orientation and $\boldsymbol{\theta} \in \mathbb{R}^{6K}$ denotes the rotations with $K$ joints. Then the pose encoder $\mathcal{E}$ that consists of several 1-D convolutional layers projects  $\boldsymbol{p}$ into a latent embedding  $\boldsymbol{z} = \mathcal{E}(\theta)$ with $\boldsymbol{z}\in \mathbb{R}^{L_p\times d_p}$, where $L_p$ is the number of pose tokens and $d_p$ is the latent dimension. Next, we transform  $\boldsymbol{z}$ into a collection of codebook entries through discrete quantization. Specifically, the  process of quantization replaces each item of $\boldsymbol{z}$ with its nearest entry in the codebook $\mathcal{B}_p$, obtaining the quantized latent vector $\widehat{\boldsymbol{z}} \in \mathbb{R}^{L_p\times d_p}$ as follows:
\begin{equation}
    \widehat{\boldsymbol{z}} = \underset{b_m \in \mathcal{B}_p}{\arg\min}\left\|\boldsymbol{z} - b_m\right\|_2. 
    \vspace{-2mm}
\end{equation}After quantization, the pose decoder $\mathcal{D}$, consisting of several 1-D deconvolutional layers, projects  $\widehat{\boldsymbol{z}}$  back to raw pose space as $\widehat{\boldsymbol{p}} = \mathcal{D}(\widehat{\boldsymbol{z}})$.
Following \cite{van2017vqvae}, we train the pose tokenizer using the loss function $\mathcal{L}_{vq}  = \mathcal{L}_{r} + \mathcal{L}_e + \mathcal{L}_c$ where $\mathcal{L}_r$, $\mathcal{L}_e$, and $\mathcal{L}_c$ denote reconstruction loss, embedding loss and commitment loss respectively. Further training and objective details are provided in the Appendix.

After training the pose tokenizer, the pose $\boldsymbol{p}$ can be represented as a sequence of discrete codebook indices of the quantized latent vector, namely \textit{pose tokens} $\mathbf{u} \in \mathbb{R}^{L_p}$ as:
\begin{equation}
    \mathbf{u} = \underset{m \in \left\{1,\dots,M\right\}}{\arg\min}\left\|\boldsymbol{z} - b_m\right\|_2. 
\end{equation}
\vspace{-4mm}
\subsection{Pose-aware Vision-Language Model}\label{pose-vlm}
\vspace{-1mm}

\noindent\textbf{Visual Processor.} \
% To enable pose comprehension from visual inputs, we equip UniPose with a visual processor that encodes image as a sequence of visual embeddings. 
Previous works \cite{liu2023llava, bai2023qwen-vl} commonly use CLIP visual encoder \cite{radford2021clip}  as the visual branch. However, since CLIP is optimized by global and coarse-grained supervision signals from image captions, it struggles to capture pose-relevant details. 
Differently, the pose estimation task demands precise localization of human keypoints, which encourages the visual encoder to capture fine-grained pose features.
Therefore,  we integrate a pose-specific Vision Transformer \cite{goel2023hmr2.0}, pretrained on the pose estimation task, into the visual branch, as shown in Fig. \ref{fig:framework}. 
Specifically, denote the CLIP visual encoder and pose-specific vision transformer as $f_a$ and $f_b$, respectively. Given an input image $\boldsymbol{x}$, we extract visual embeddings by CLIP  as $\mathbf{v_a}=f_a\left(\boldsymbol{x}\right)$ where $\mathbf{v_a}\in\mathbb{R}^{L_v \times d_a}$, $L_v$ is the number of visual patch tokens and $d_a$ is its visual embedding dimension. The embedding output by pose-specific vision transformer is $\mathbf{v_b}=f_b\left(\boldsymbol{x}\right)$ where $\mathbf{v_b}\in\mathbb{R}^{L_v \times d_b}$. 
Then we concatenate the embedding output by these two encoders along the channel dimension, and apply a trainable projector layer (with projection matrix $W\in\mathbb{R}^{(d_a + d_b) \times d}$) to align the dimension of the concatenated visual features to that of text features as $\mathbf{v}=\left[\mathbf{v_a} | \mathbf{v_b}\right]^TW$. Here  $\mathbf{v}\in\mathbb{R}^{L_v \times d}$ with $d$ as text embedding dimensions of LLM.
The fused visual features $\mathbf{v}$ can be concatenated with pose or text tokens as input to LLM.

\noindent \textbf{Mixed Attention Mechanism.} \
Existing LLMs \cite{raffel2020t5, jiang2023mistral, touvron2023llama} typically employ autoregressive modeling with causal attention, excelling at generating sequential data such as text and audio \cite{yang2024uniaudio, zhang2023speechgpt}. However, pose tokens, which encode spatial positions of human joints,  are inherently non-sequential, making traditional autoregressive generation suboptimal. To address this issue, we propose modeling pose tokens as a whole. Inspired by \cite{xie2024showo,zhou2024transfusion}, we modify the standard causal attention in LLM, integrating bidirectional attention for pose tokens as depicted in Fig. \ref{fig:framework}. Specifically, we apply casual attention to text sequence, but apply bidirectional attention within the pose token sequence. To avoid information leakage, we initialize $L_p$ learnable pose queries $\mathcal{Q}=\{q_1,...,q_{L_p}\}$ during the generation and editing of 3D poses, as shown in Fig. \ref{fig:framework}. These queries are used to predict corresponding pose tokens in a single forward step. This design enables each pose token to attend to others within the same pose token sequence, while restricting access to only previously encountered text tokens.

\begin{figure*}[t]
    \centering
    \begin{subfigure}{0.32\linewidth} 
        \centering
        \includegraphics[width=0.95\linewidth]{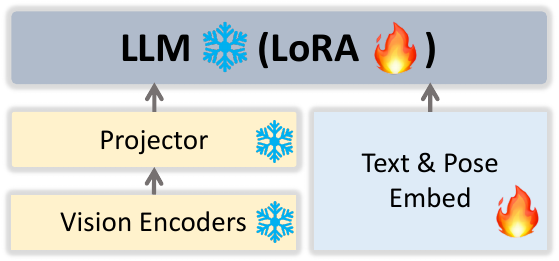}
        \caption{Pose-Text Alignment Pretraining Stage. }
        \label{fig:stage2}
    \end{subfigure}
    \begin{subfigure}{0.32\linewidth}  
        \centering
        \includegraphics[width=0.95\linewidth]{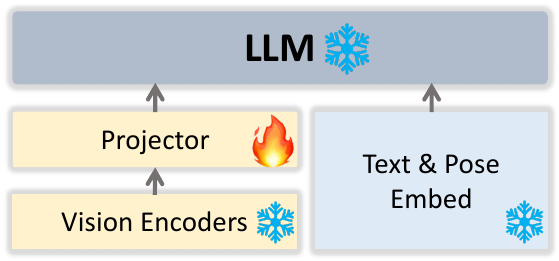}
        \caption{Visual Projector Pretraining Stage. }
        \label{fig:stage3}
    \end{subfigure}
    \begin{subfigure}{0.32\linewidth} 
        \centering
        \includegraphics[width=0.95\linewidth]{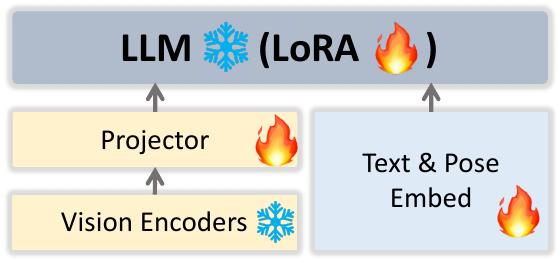}
        \caption{Instruction Finetuning Stage. }
        \label{fig:stage4}
    \end{subfigure}
    \caption{The training paradigm of UniPose.}
    \vspace{-4mm}
    \label{fig:stage}
\end{figure*}

\noindent \textbf{Unified Multimodal Language Model.} \
As shown in Fig.~\ref{fig:framework}, equipped with a visual processor and pose tokenizer, we can compress original visual data $\boldsymbol{x}$ and pose data $\boldsymbol{p}$ into visual feature sequence $\mathbf{v}\in\mathbb{R}^{L_v \times d}$ and pose token sequence $\mathbf{u} \in \mathbb{R}^{L_p}$, respectively. \textbf{}
To incorporate pose discrete tokens into LLMs, we expand the original text vocabulary $\mathcal{V}_t$ of LLM with pose vocabulary $\mathcal{V}_p$\footnote{The pose vocabulary $\mathcal{V}_p$ preserves the order of the pose codebook $\mathcal{B}_p$. In implementation, we add two special tokens, {\ttfamily <p>} and {\ttfamily <p/>}, which denotes the start and end  of a pose sequence, into the vocabulary $\mathcal{V}_p$.}, forming a new unified text-pose vocabulary $\mathcal{V} = \{\mathcal{V}_t, \mathcal{V}_p\}$.
Equipped with the unified vocabulary $\mathcal{V}$, various pose-related tasks can be formulated in a general format, where both input and output tokens are drawn from the same vocabulary, with the input optionally combined with the visual feature $\mathbf{v}$. 
These discrete tokens can represent natural language, 3D pose, or combination, depending on the specific task to be solved. 
This naturally enables  UniPose to unify pose comprehension, generation, and editing in a unified manner.

During training, denote the visual embedding sequence as $\mathbf{v}=\left\{v^i \in\mathbb{R}^d \right\}_{i=1}^{L_v}$, the pose token sequence as $\mathbf{u} = \left\{u^i \in \mathcal{V} \right\}_{i=1}^{L_p}$, the text token sequence of single-pose  description as  $\mathbf{t}=\left\{t^i \in \mathcal{V} \right\}_{i=1}^{L_t}$, and the text token sequence of pose-difference description as $\mathbf{d}=\left\{d^i \in \mathcal{V} \right\}_{i=1}^{L_d}$,  we apply distinct optimization objectives  for each task, tailored to the specific type of input and the desired output, as follows:

\begin{itemize}
	\item \textit{Single-Pose Comprehension.} Single-pose comprehension aims to generate a pose description from a 3D pose or image. Formally, given the sequence $\mathbf{v}$,  $\mathbf{u}$ and  $\mathbf{t}$ as defined above, the LLM predicts the probability distribution of potential next text token at each step, $p_{\theta}\left(t^i| \mathbf{v}/\mathbf{u}, t^{<i}\right)$,   conditioned on the visual or pose tokens in an autoregressive manner. The objective is to maximize the log-likelihood of this predicted pose description distribution: 
	\begin{equation}
    \vspace{-2mm}
    \mathcal{L}_{1} = \sum_{i=1}^{L_t}\log p_{\theta}\left(t^i| \mathbf{v}/\mathbf{u}, t^{<i}\right),
	\end{equation}
	where $\theta$ represents the trainable parameters.

	\item \textit{Pose-pair Comprehension.} Pose-pair comprehension aims to generate a textural description of the difference between a pair of 3D poses or images. Formally, given visual features $\mathbf{v_1}$ and $\mathbf{v}_2$ for an image pair, pose tokens $\mathbf{u_1}$ and $\mathbf{u_2}$ for a 3D pose pair, and pose-difference description tokens $\mathbf{d}$, the LLM predicts the probability distribution of the next pose-difference text token, $p_{\theta}\left(d^i| \left( \mathbf{v_1}, \mathbf{v_2}\right)/\left(\mathbf{u_1}, \mathbf{u_2}\right), d^{<i}\right)$, conditioned on the pair of visual or pose tokens in an autoregressive manner. The objective is to maximize the log-likelihood of this predicted pose-difference description distribution:  
	\begin{equation}
    \mathcal{L}_{2} = \sum_{i=1}^{L_d}\log p_{\theta}\left(d^i| \left( \mathbf{v_1}, \mathbf{v_2}\right)/\left(\mathbf{u_1}, \mathbf{u_2}\right), d^{<i}\right).
	\end{equation}

	\item \textit{Pose Generation.} Pose generation aims to generate 3D poses from pose textural descriptions or images. For this task, we use a \textit{mixed attention mechanism} where the input pose tokens are replaced with  predefined pose queries $\mathcal{Q}$. %in the input. 
    Formally, given  $\mathbf{v}$, $\mathbf{t}$ and $\mathbf{u}$ as defined above, LLM predicts the probability distribution of potential \textit{whole} pose tokens in a single step, $p_{\theta}\left(\mathbf{u} | \mathbf{v}/\mathbf{t}, \mathcal{Q}\right)$, conditioned on the visual or pose-description text tokens and pose queries. The objective is to maximize  the log-likelihood of this predicted pose distribution:   
	\begin{equation}
    \mathcal{L}_{3} = p_{\theta}\left(\mathbf{u} | \mathbf{v}/\mathbf{t}, \mathcal{Q}\right).
	\end{equation} 

	\item \textit{Pose editing.} Pose editing aims to generate a corrected 3D pose based on an initial pose and modification instruction. Similar to pose generation, a \textit{mixed attention mechanism} is used for this task.  Formally, given $\mathbf{u_1}$, $\mathbf{u_2}$ and $\mathbf{d}$ as defined above, LLM predicts the probability distribution of potential \textit{whole} pose tokens for the corrected pose, $p_{\theta}\left(\mathbf{u_2} | \mathbf{d}, \mathbf{u_1}, \mathcal{Q}\right)$,  conditioned on the initial pose tokens, modification instruction tokens and pose queries. The objective is to maximize  the log-likelihood of this predicted corrected-pose distribution:   
	\begin{equation}
    \mathcal{L}_{4} = p_{\theta}\left(\mathbf{u_2} | \mathbf{u_1}, \mathbf{d}, \mathcal{Q}\right).
	\end{equation}
\end{itemize}
\vspace{-2mm}
At the last, given a batch size of inputs with different task types, the overall training loss is computed as the sum of the individual objections: $\mathcal{L}=\mathcal{L}_1+\mathcal{L}_2+\mathcal{L}_3+\mathcal{L}_4$. 
% \begin{equation}
% \mathcal{L}=\mathcal{L}_1+\mathcal{L}_2+\mathcal{L}_3+\mathcal{L}_4.
%   \label{eq8}
% \end{equation}

\begin{table*}[]
\centering
\footnotesize
\begin{tabular}{l|l|l|ccc|ccc}
\toprule
\multirow{2}{*}{Task} & \multirow{2}{*}{Dataset} & \multirow{2}{*}{Method} & \multicolumn{3}{c|}{R-Precision $\uparrow$} & \multicolumn{3}{c}{Linguistic metrics $\uparrow$} \\ \cmidrule{4-9} 
 & & & Top-1 & Top-2 & Top-3 & BLEU-4  & ROUGE-L & METEOR \\
\midrule
\multirow{3}{*}{Pose-to-Text} & \multirow{3}{*}{PoseScript \cite{delmas2024posescript}} & PoseScript \cite{delmas2024posescript} & \textbf{91.6} & \textbf{95.6} & 97.0 & \textbf{12.9} & \textbf{33.9} & \textbf{34.2} \\
& & UniPose $\dagger$ & 18.1 & 30.0 & 39.1 & 10.8 & 30.1 & 29.5 \\
& & UniPose & 85.6 & 95.2 & \textbf{97.6} & 12.1 & 33.3 & 30.8 \\
\midrule
\multirow{3}{*}{Pose-Diff} & \multirow{3}{*}{PoseFix \cite{delmas2023posefix}} & PoseFix \cite{delmas2023posefix} & 64.6 & 77.1 & 83.0 & 12.0 & 33.5 & \textbf{36.7} \\
& & UniPose $\dagger$& 8.4 & 14.6 & 19.2 & 8.5 & 28.2 & 27.3 \\
& & UniPose & \textbf{67.9} & \textbf{81.8} & \textbf{88.6} & \textbf{13.8} & \textbf{33.7} & 31.2 \\
\midrule
\multirow{5}{*}{Image-to-Text} & \multirow{5}{*}{ImageScript} & LLaVA \cite{liu2023llava} & 5.7 & 12.0 & 18.9 & 3.2 & 21.8 & 32.9 \\
& & Qwen-VL \cite{bai2023qwen-vl} & 8.9 & 15.6 & 19.8 & 1.4 & 15.9 & 21.6 \\
& & GPT4V \cite{2023arXivgpt4} & 17.7 & 24.0 & 32.3 & 7.1 & 29.1 & 34.2 \\
& & UniPose $\dagger$ & 22.4 & 32.8 & 41.2 & \textbf{18.2} & 42.4 & \textbf{45.2} \\
& & UniPose & \textbf{24.5} & \textbf{35.4} & \textbf{43.2} & \textbf{18.2} & \textbf{42.5} & 44.7 \\
\midrule
\multirow{3}{*}{Image-Diff} & \multirow{3}{*}{ImageDiff} & GPT4V \cite{2023arXivgpt4} & 7.3 & 13.5 & 18.8 & 1.3 & 16.1 & 21.8 \\
& & UniPose $\dagger$ & 13.0 & 18.8 & 26.4 & 14.0 & 34.1 & 40.1 \\
& & UniPose & \textbf{13.5} & \textbf{25.0} & \textbf{33.8} & \textbf{15.9} & \textbf{36.5} & \textbf{39.6} \\
\bottomrule
\end{tabular}
\caption{\textbf{Comparisons on pose comprehension tasks}. We compare the pose-retrieval precision (R-Precision) and linguistic metrics on various datasets. %For Image-to-Text and Image-Diff tasks, we calculate the R-Precision using the predicted text and the ground truth SMPL pose of the image. Models marked with $\dagger$ are trained on single tasks.
UniPose $\dagger$ represents training UniPose on the single corresponding task.
%the model trained on a single corresponding task. %pose comprehension task.
}
\label{tab-pose-understanding}
\vspace{-5mm}
\end{table*}

\subsection{Training and Inference Paradigm}
\label{training-strategy}

The training procedure comprises four stages, and  the training paradigm of the last three stages is shown in Fig. \ref{fig:stage}.
% The training process is divided into four stages:  
% %\textbf{\romannumeral 1)} 
% (1) Pose Tokenizer Training: learning the pose tokenizer to represent 3D pose as discrete tokens. 
% %\textbf{\romannumeral 2)}
% (2) Pose-Text Alignment Pretraining: aligning the pose and text modalities.  The goal is to enable the model to understand the correspondence between 3D poses and their textual descriptions, establishing a shared space between these two modalities.
% %\textbf{\romannumeral 3)} 
% (3) Visual Projector Pretraining: aligning the visual and text modalities. By integrating visual features with textual descriptions, this stage helps the model understand and generate pose-related information from visual inputs, such as images.
% %\textbf{\romannumeral 4)} 
% (4) Instruction Finetuning:  enhancing the model’s instruction-following capability. 
% The training paradigm of the last three stages is shown in Fig. \ref{fig:stage}.

\noindent
\textbf{Pose Tokenizer Training.} 
 We first train a pose tokenizer using the objective $\mathcal{L}_{vq}$. The pose tokenizer encodes 3D pose as a sequence of discrete tokens, enabling seamless integration with texts within LLM. To maintain stability during LLM training, the pose tokenizer is kept frozen during the subsequent stages of training. 

 \noindent\textbf{Pose-Text Alignment Pretraining.}
To enable LLM to handle discrete pose tokens, we train LLM on pose-text corpus. This process aims to align the pose and text modalities for unified reasoning within the LLM. 
In this stage, we consider four pose-text relevant tasks in Tab.~\ref{tab-pose-tasks}, \textit{i.e.}, $2$ pose comprehension tasks (Pose-to-Text and Pose-Diff), $1$ pose generation task (Text-to-Pose) and the Pose Editing task.
Based on these tasks, we train LLM using  LoRA \cite{hu2021lora} with the objective $\mathcal{L}$, as shown in Fig. \ref{fig:stage2}.

\noindent\textbf{Visual Projector Pretraining.}
 After establishing alignment between pose and text modalities, this training stage focuses on mapping images into the shared pose-text space. In this stage, we consider three image-text relevant tasks in Tab.~\ref{tab-pose-tasks}, \textit{i.e.}, $2$ pose comprehension tasks (Image-to-Text and Image-Diff) and $1$ pose generation task (Image-to-Pose). Based on these tasks, we train the vision-language projector to align visual data with language models with the objective $\mathcal{L}$, as shown in Fig. \ref{fig:stage3}.

\noindent\textbf{Instruction Finetuning. } To enhance the instruction-following capability of UniPose, we construct a multitask, multimodal instruction dataset with 200 templates for each task from Tab.~\ref{tab-pose-tasks}. For example, an instruction for Image-to-Pose task could be ``{\ttfamily Could you estimate the SMPL pose of the individual in this image} {\ttfamily <image>}'', with {\ttfamily <image>} standing for the image embedding extracted by the visual processor. Using this instruction data, we jointly train the visual projector and LLM with LoRA, as shown in Fig. \ref{fig:stage4}.

\noindent \textbf{Inference. }
% \subsection{Inference}
% \label{sec:inference}
During inference, we adopt tailored decoding strategies according to task type. For pose comprehension tasks, we use a standard auto-regressive approach, where text tokens are generated sequentially, step-by-step. For pose generation and editing tasks, as shown in Fig. \ref{fig:framework}, once the model predicts the $\mathrm{start\_of\_pose}$ token {\ttfamily <p>}, we append $L_p$ predefined pose queries to the conditional text tokens, which is fed into LLM. Then LLM predicts the corresponding pose token for each query in parallel, which significantly accelerates its inference speed. 
% More details can be found in the Appendix. 

\section{Experiments}
\label{experiments}
\subsection{Experimental Setup}

\begin{table*}[]
\centering
\footnotesize
% \resizebox{0.98\linewidth}{!}{
\begin{tabular}{l|ccc|ccc|ccc}
\toprule
\multirow{2}{*}{Method} & \multicolumn{3}{c|}{$\mathrm{R}^{\mathrm{T2P}} \uparrow$} & \multicolumn{3}{c|}{$\mathrm{R}^{\mathrm{P2T}} \uparrow$} & \multicolumn{3}{c}{Pose Reconstruction Metric $\downarrow$} \\ \cmidrule{2-10} 
 & Top-5 & Top-10 & Top-20 & Top-5 & Top-10 & Top-20 & MPJPE & PA-MPJPE & FID \\
\midrule
PoseScript \cite{delmas2024posescript} & 73.3 & \textbf{82.5} & 89.4 & 70.0 & \textbf{82.5} & 87.4 & 318.0 & \textbf{161.3} & 0.075 \\
ChatPose \cite{feng2024chatpose} & 17.6 & 25.3 & 35.8 & 28.0 & 39.0 & 54.4 & - & - & - \\
ChatHuman \cite{lin2024chathuman} & 41.8 & 52.6 & 65.1 & 42.1 & 52.3 & 66.5 & - & - & - \\
\hline
UniPose $\dagger$ & 67.5 & 77.6 & 85.5 & 62.8 & 74.8 & 83.6 & 342.7 & 190.0 & 0.046\\
UniPose & \textbf{73.7} & 82.4 & \textbf{89.6} & \textbf{70.9} & 80.5 & \textbf{89.6} & \textbf{308.6} & 171.1 & \textbf{0.038} \\
\bottomrule
\end{tabular}
% }
\caption{\textbf{Comparisons on Text-to-Pose generation task}. The retrieval and reconstruction metrics are reported on PoseScript \cite{delmas2024posescript} dataset. 
% UniPose $\dagger$ denotes the model trained on single text-to-pose generation task.
}
\label{tab-text2pose}
\vspace{-4mm}
\end{table*}

\begin{table}[]
\centering
\footnotesize
\resizebox{0.98\linewidth}{!}{
\begin{tabular}{l|c|c|c|c}
\toprule
\multirow{2}{*}{Method} & \multicolumn{2}{c|}{3DPW \cite{von2018-3dpw} $\downarrow$ } & \multicolumn{2}{c}{H3.6M \cite{ionescu2013human3.6m} $\downarrow$ } \\ \cmidrule{2-5} 
 & MPJPE & PA-MPJPE & MPJPE & PA-MPJPE  \\
\midrule
HMR \cite{kanazawa2018hmr} & 130.0 & 76.7 & 88.0 & 56.8 \\
PyMAF \cite{pymaf2021} & 92.8 & 58.9 & 57.7 & 40.5 \\
SMPLer \cite{xu2023smpler} & 73.7 & 43.4 & 45.2 & \textbf{32.4} \\
HMR2.0 \cite{goel2023hmr2.0} & 70.0 & 44.5 & \textbf{44.8} & 33.6 \\
Zolly \cite{wang2023zolly} & 76.2 & 47.9 & 49.4 & 32.3 \\
MEGA \cite{fiche2024mega} & \textbf{67.5} & \textbf{41.0} & - & - \\
TokenHMR \cite{dwivedi2024tokenhmr} & 71.0 & 44.3 & - & - \\
\midrule
ChatPose \cite{feng2024chatpose} & 163.6 & 81.9 & 126.0 & 82.4 \\
UniPose $\dagger$ & 97.4 & 61.2 & \textbf{65.8} & \textbf{39.4} \\
UniPose & \textbf{94.7} & \textbf{59.1} & 69.2 & 41.8 \\
\bottomrule
\end{tabular}}
\caption{\textbf{Comparisons on pose estimation task.} Reconstruction metrics are reported on the 3DPW and Human3.6M datasets.
% UniPose $\dagger$ denotes the model trained on single pose estimation task.
}
\label{tab-image2pose}
\vspace{-0.5cm}
\end{table}

\noindent\textbf{Datasets}. For pose tokenizer training, we use the standard training split of AMASS \cite{mahmood2019amass} and MOYO \cite{tripathi2023moyo}, following TokenHMR \cite{dwivedi2024tokenhmr}. 
For UniPose training, we integrate three types of data: 
%\textbf{\romannumeral 1) 
\textbf{(1)} Text-Pose Data. We use PoseScript \cite{delmas2024posescript} and PoseFix \cite{delmas2023posefix} datasets to link language and pose modality. PoseScript \cite{delmas2024posescript} provides natural language descriptions paired with 3D human poses, allowing the model to understand fine-grained pose semantics. PoseFix \cite{delmas2023posefix} includes pairs of 3D poses and textual descriptions that specify how to modify the source pose to achieve the target pose. 
%\textbf{\romannumeral 2) 
\textbf{(2)} Image-Pose Data. Following  \cite{dwivedi2024tokenhmr, goel2023hmr2.0}, we use standard human pose estimation training datasets, including Human3.6M \cite{ionescu2013human3.6m}, MPI-INF-3DHP \cite{mehta2017mpi-inf-3dhp}, COCO \cite{lin2014coco}, and the MPII \cite{andriluka20142d-mpii} dataset, and evaluate on 3DPW \cite{von2018-3dpw} and Human3.6M \cite{ionescu2013human3.6m} test sets. %Note that we omit estimating the SMPL shape parameters, as they are not directly related to human pose. 
%\textbf{\romannumeral 3) 
\textbf{(3)} Image-Text Data. Since no existing dataset combines human images with pose descriptions, we create the ImageScript and ImageDiff datasets to bridge this gap in visual-textual pose comprehension. Further dataset details are provided in the Appendix.

\noindent\textbf{Metrics}. We adopt the evaluation metrics from PoseScript \cite{delmas2024posescript} and PoseFix \cite{delmas2023posefix}. %\textbf{\romannumeral 1) P
\textbf{(1) Pose comprehension tasks.} We use two types of metrics.  Pose-text retrieval metric: \textit{R-Precision}, which evaluates the accuracy of matching poses with corresponding descriptions. We rank the Euclidean distances between the query pose and 32 text descriptions ($1$ ground truth and $31$ randomly selected mismatched descriptions), and report Top $1 / 2 / 3$ R-Precision; Linguistic metrics: \textit{BLEU-4} \cite{papineni2002bleu}, \textit{Rouge-L} \cite{lin2004rouge} and \textit{METEOR} \cite{banerjee2005meteor}, which assess the quality of generated pose descriptions. 
%\textbf{\romannumeral 
\textbf{(2) Pose generation tasks.} We use two types of metrics. Reconstruction metrics: \textit{MPJPE} and \textit{PA-MPJPE}, which computes the average per-joint position error between generated and ground-truth pose; Pose-text retrieval metric: following \cite{feng2024chatpose}, we report Top $5 / 10 / 20$ $\mathrm{R}^{\mathrm{T2P}}$ and $\mathrm{R}^{\mathrm{P2T}}$, which represents the text-to-pose and pose-to-text retrieval recall, respectively. 
%\textbf{\romannumeral 3) 
\textbf{(3) Pose editing tasks.} In addition to the reconstruction metrics, we also report the \textit{Frechet Inception Distance (FID)}, which measures the distance between the generated and ground-truth pose distribution.
To calculate these metrics, following \cite{delmas2024posescript, delmas2023posefix}, we train a retrieval model with pose and text feature extractors using contrastive loss, which encourages matched pose-text pairs to have geometrically close feature vectors.

\noindent\textbf{Implementation Details}.
For pose tokenizer, we set the codebook size to $2048$ and each 3D pose is represented with $80$ discrete tokens. We utilize  LLaVA-1.6V \cite{liu2023llava} as our visual-language model backbone.
For training the pose tokenizer, we use AdamW as the optimizer with a batch size of 256 and an initial learning rate of 2e-4. 
% To train the multimodal backbone, we employ AdamW optimizer. 
% For hyperparameter setting, $\lambda_1$, $\lambda_2$ and $\lambda_3$ (Eq.~\ref{eq2}) are set to $20$, $100$, $100$ respectively. 
The pose tokenizer is trained for $240$ epochs on a single RTX 4090 GPU. UniPose is trained $6$ epochs in the Pose-Text Alignment Pretraining stage, and $2$ epochs in the remaining stages using 4 A100 GPUs.
Further implementation details are provided in the Appendix.

\begin{table*}
\begin{minipage}[t]{0.4\textwidth}
\centering
\footnotesize
    \renewcommand{\arraystretch}{1.15}
    \begin{tabular}{l|ccc}
    \toprule
    Method & MPJPE $\downarrow$ & PA-MPJPE $\downarrow$ & FID $\downarrow$ \\
    \midrule
    PoseFix \cite{delmas2023posefix} & 300.2 & 144.1 & 0.019 \\
    UniPose $\dagger$ & 310.8 & 157.0 & 0.019 \\
    UniPose & \textbf{270.3} & \textbf{138.9} & \textbf{0.015} \\
    \bottomrule
    \end{tabular}
    \caption{\textbf{Comparisons on pose editing task}. Reconstruction metrics are reported on PoseFix \cite{delmas2023posefix} dataset.
    % UniPose $\dagger$ denotes the model trained on single pose editing task.
    }
    \renewcommand{\arraystretch}{1}
    \label{tab-pose-edit}
\end{minipage}
\begin{minipage}[t]{0.6\textwidth}
\centering
\footnotesize
\begin{tabular}{cc|cc|ccc}
\toprule
 \multirow{2}{*}{CLIP-ViT} & \multirow{2}{*}{Pose-ViT} & \multicolumn{2}{c|}{Pose Estimation $\downarrow$} & \multicolumn{3}{c}{Image-to-Text $\uparrow$} \\ \cmidrule{3-7}
 & & MPJPE & PA-MPJPE & BLEU-4 & ROUGE-L & METEOR \\ 
\midrule
\ding{52} & \ding{56} & 193.4 & 86.1 & 11.1 & 30.2 & 33.9 \\
\ding{56} & \ding{52} & 96.3 & 59.1 & 12.5 & 31.0 & 34.8 \\
\ding{52} & \ding{52} & \textbf{96.1} & \textbf{58.9} & \textbf{13.3} & \textbf{31.7} & \textbf{35.2} \\
\bottomrule
\end{tabular}
\caption{\textbf{Ablation study on the components of the visual processor. }}
\label{tab:visual}
\end{minipage}
\end{table*}

\begin{table*}[]
\centering
\footnotesize
\begin{tabular}{l|ccc|ccc|c|ccc}
\toprule
\multirow{2}{*}{Attention Type} & \multicolumn{7}{c|}{Text-to-Pose} & \multicolumn{3}{c}{Pose-to-Text } \\ \cmidrule{2-11}
& \multicolumn{3}{c}{$\mathrm{R}^{\mathrm{T2P}} \uparrow$} & \multicolumn{3}{|c|}{$\mathrm{R}^{\mathrm{P2T}} \uparrow$} & Latency (s) $\downarrow$ & BLEU-4 $\uparrow$ & ROUGE-L $\uparrow$ & METEOR $\uparrow$\\
\midrule
Causal Attention & 9.0 & 14.2 & 20.8 & 9.3 & 14.7 & 22.3 & 2.5 & \textbf{26.9} & \textbf{39.5} & \textbf{38.0} \\
Mixed Attention & \textbf{13.8} & \textbf{20.3} & \textbf{28.8} & \textbf{15.9} & \textbf{23.0} & \textbf{32.0} & \textbf{0.2} & 25.0 & 39.1 & 36.7 \\
\bottomrule
\end{tabular}
\caption{\textbf{Ablation study on different attention mechanisms.} %Top 5 / 10 / 20 retrieval recall rates are reported on the PoseScript dataset.}
}
\label{tab:mask}
\vspace{-6mm}
\end{table*}

\subsection{Comparisons on Pose-relevant Tasks}

\noindent\textbf{Comparisons on Pose comprehension.} We evaluate UniPose on $4$ pose comprehension tasks, \textit{i.e.}, Pose-to-Text, Pose-Diff, Image-to-Text and Image-Diff. The comparison results are shown in Tab. \ref{tab-pose-understanding}. As seen in the table, UniPose achieves competitive performance across all evaluated  tasks, highlighting its capability to address diverse pose comprehension tasks within a single model. (1) For Pose-to-Text task, we compare UniPose with PoseScript \cite{delmas2024posescript} on the PoseScript dataset.  As shown in Tab. \ref{tab-pose-understanding}, UniPose achieves slightly lower performance than PoseScript. However, PoseScript is tailored for single-pose description generation and lacks the capacity to model relationships between different poses. 
(2) For Pose-Diff task, we compare UniPose with PoseFix \cite{delmas2023posefix} on the PoseFix dataset. As shown in  Tab. \ref{tab-pose-understanding}, UniPose outperforms PoseFix on most metrics, demonstrating its superiority in capturing relationships between pairs of poses.
(3) For Image-to-Text task, we compare UniPose with existing visual-language MLLMs, including   LLaVA \cite{liu2023llava}, Qwen-VL \cite{bai2023qwen-vl} and GPT4V \cite{2023arXivgpt4}, on the ImageScript dataset. As shown in  Tab. \ref{tab-pose-understanding}, UniPose significantly outperforms these MLLMS. The substantial gains can be attributed to the use of pose-specific visual encoder, which enables UniPose to extract fine-grained pose information from visual inputs. %, thereby enhancing UniPose’s ability to accurately perceive and understand poses from images.
(4) For Image-Diff task, we compare UniPose with GPT4V on the ImageDiff dataset. UniPose still outperforms GPT4V, demonstrating that UniPose not only captures fine-grained pose features from a single image but also learns the relationships between human poses across multiple images. 

\begin{figure}[t]
    \centering
    \begin{subfigure}{\linewidth}  % 第一行第一个子图
         \centering
         \includegraphics[width=\linewidth]{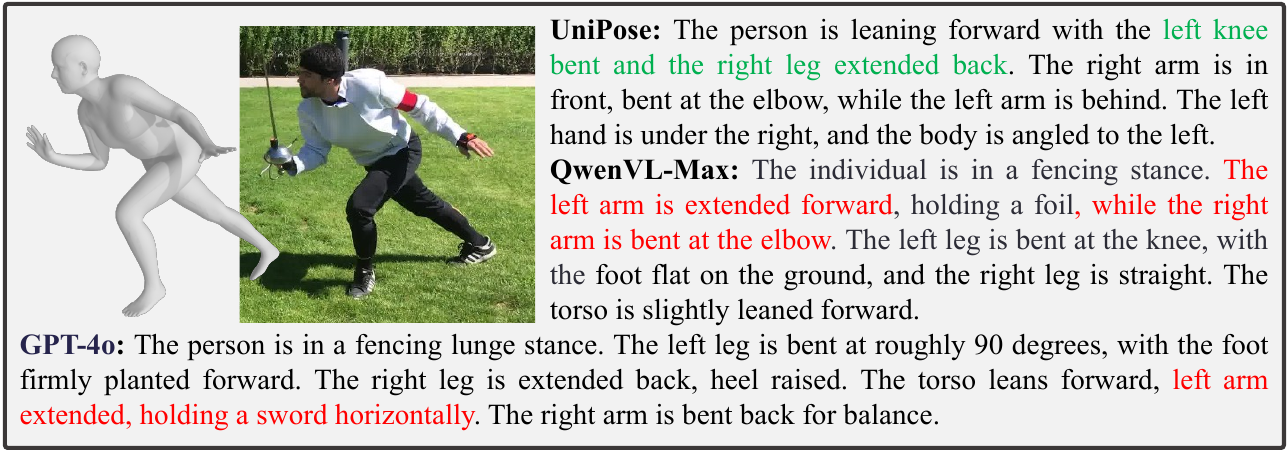}
         \caption{Comparison with Qwen-VL \cite{bai2023qwen-vl} and GPT-4o \cite{2023arXivgpt4} on Image-to-Text task. }
         \label{fig:image2text-vis}
    \end{subfigure}
    
    \begin{subfigure}{\linewidth}  % 第一行第二个子图
         \centering
         \includegraphics[width=\linewidth]{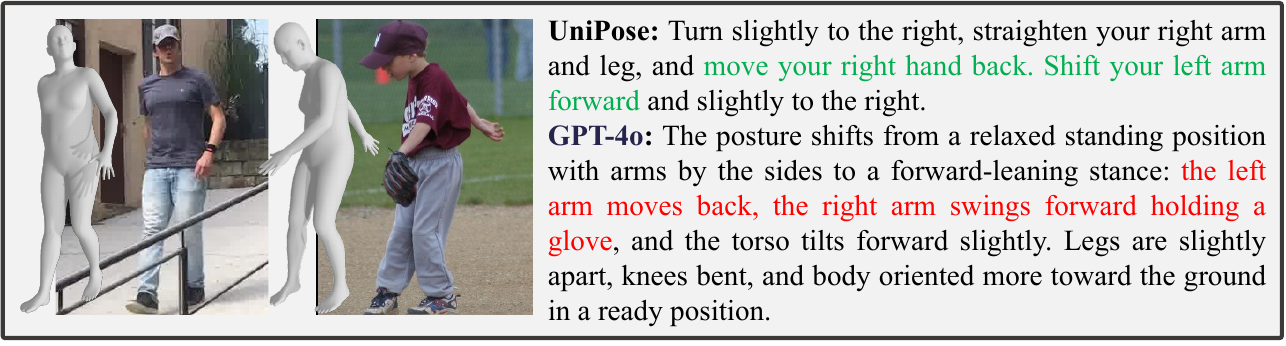}
         \caption{Comparison with GPT-4o \cite{2023arXivgpt4} on Image-Diff task. }
         \label{fig:image-diff-vis} 
    \end{subfigure}
    \caption{\textbf{Examples on Image-to-Text and Image-Diff tasks}. We mark incorrect captions in \textcolor{red}{red} and correct in \textcolor{ForestGreen}{green}. UniPose can accurately perceive a person's orientation from images.}
    \label{fig:vis}
    \vspace{-0.7cm}
\end{figure}

Fig. \ref{fig:image2text-vis} and Fig. \ref{fig:image-diff-vis} visualize the generated textural descriptions of UniPose, Qwen-VL \cite{bai2023qwen-vl} and GPT4V \cite{2023arXivgpt4}. The visualizations reveal that existing MLLMs struggle to comprehend fine-grained pose information. Specifically, Qwen-VL \cite{bai2023qwen-vl} and GPT4V \cite{2023arXivgpt4} fail to distinguish human body orientation, whereas UniPose can accurately locate the human body orientation from visual inputs.

\noindent\textbf{Comparisons on Pose Generation.}  We further evaluate UniPose on $2$ pose generation tasks, \textit{i.e.}, text-to-pose and pose estimation.  The comparison results are shown in Tab. \ref{tab-text2pose} and Tab. \ref{tab-image2pose}. 
(1) For Text-to-Pose task, we compare UniPose with existing text-conditional pose generation models \cite{delmas2024posescript, feng2024chatpose, lin2024chathuman} on PoseScript dataset. As shown in  Tab. \ref{tab-text2pose}, UniPose achieves the best performance on most metrics.  We attribute this to the use of the mixed-attention mechanism in LLM, which effectively captures the bidirectional dependencies among pose tokens, thus improving pose generation performance. 
(2) For pose estimation task, we compare UniPose with traditional pose estimation approaches \cite{dwivedi2024tokenhmr, goel2023hmr2.0} and MLLM-based approaches \cite{feng2024chatpose}, on 3DPW \cite{von2018-3dpw} and H3.6M \cite{ionescu2013human3.6m} datasets. As shown in  Tab. \ref{tab-image2pose}, UniPose largely outperforms other MLLMs, yet does not match the performance of methods specifically designed for estimating 3D human pose. This is not surprising, as traditional pose estimation methods have been developed over many year and often incorporate custom network modules and loss functions to enhance estimation accuracy.   

% Figure xx provides visualizations of pose estimation results.

\noindent\textbf{Comparisons on Pose Editing.} For pose editing task, we compare UniPose with PoseFix \cite{delmas2023posefix} on PoseFix dataset. As shown in Tab. \ref{tab-pose-edit}, UniPose significantly outperforms PoseFix  across all metrics, validating its superiority in pose editing. 
\vspace{-4mm}

\subsection{Ablation Studies \& Analysis}

\noindent\textbf{Single-task training \textit{v.s.} Multi-task training.} 
Tab. \ref{tab-pose-understanding}, \ref{tab-text2pose}, \ref{tab-image2pose}, \ref{tab-pose-edit} also report the performance of UniPose training on single task (denoted as UniPose $\dagger$). As shown,  multi-task training consistently outperforms single-task training, underscoring the importance  of unifying pose comprehending, generation and editing within a single model. 

\noindent\textbf{Visual Processor. } We compare the impact of different vision encoders used in the Visual Processor of UniPose. In this part, the models are trained solely on Pose Estimation and Image-to-Text tasks for 2 epochs. As shown in Tab. \ref{tab:visual}, with only the CLIP-ViT encoder, the model performs poorly on pose estimation task. We argue that CLIP-ViT primarily focuses on aligning global semantic information between images and text, struggling to capture detailed human pose information. By incorporating an additional ViT model trained specifically for pose estimation, UniPose gains the ability to capture fine-grained pose details, significantly improving its performance on pose estimation task. Moreover, the pose information extracted from images enhances the performance on Image-to-Text task, enabling UniPose to generate more precise descriptions of human poses.

\noindent\textbf{Attention mechanism. } We evaluate the performance of UniPose using causal attention and mixed attention. In this part, the models are trained solely on Text-to-Pose and Pose-to-Text tasks for 6 epochs. More training details are provided in the Appendix. As shown in Tab. \ref{tab:mask}, on Text-to-Pose task, the model with mixed attention achieves higher retrieval accuracy compared to casual attention. The results indicate that capturing bidirectional dependencies among pose tokens enhances pose generation. Additionally,  the bidirectional attention mechanism enables single-step generation of all pose tokens, significantly accelerating inference. However, mixed attention performs worse than causal attention on Pose-to-Text task. This may be due to the interference of the bidirectional attention with the causal dependencies essential for text generation,  potentially compromising the semantic precision of the generated content.

%\begin{figure}[htbp]
\begin{figure}[t]
 \centering
 \includegraphics[width=\linewidth]{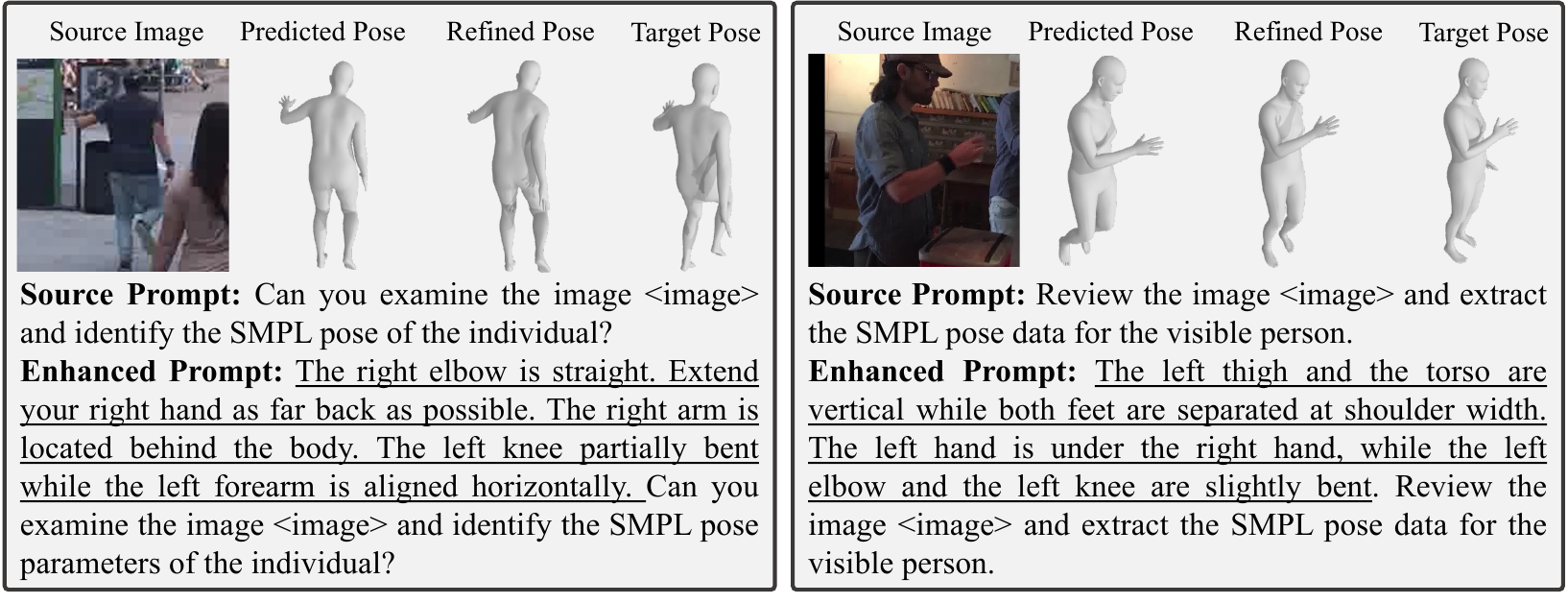}
 \caption{Enhance pose estimation with input pose description. }
 \label{fig:zero-shot}
 \vspace{-0.5cm}
\end{figure}

\noindent\textbf{Zero-shot Task Analysis. } 
Benefiting from a unified learning format, UniPose effectively transfers knowledge across different pose-relevant tasks and adapts to unseen tasks. Fig. \ref{fig:zero-shot} provides a zero-shot analysis: without additional training, UniPose can leverage pose descriptions to enhance its pose estimation results. This ability is especially advantageous in scenarios where ambiguity or occlusion affects accurate human pose estimation from images.

% UniPose demonstrates strong generalization capabilities, retaining its effectiveness even in zero-shot settings. We provide an example for zero-shot analysis: without additional training, UniPose leverages pose descriptions to enhance its performance on pose estimation tasks, as shown in Fig. \ref{fig:zero-shot}. This ability is especially advantageous in scenarios where ambiguity or occlusion affects accurate human pose estimation from images.
\vspace{-2mm}
\section{Conclusion}
\vspace{-2mm}
In this work, we present UniPose, the first attempt to integrate human pose comprehension, generation, and editing within a unified framework. By employing a pose tokenizer, we build a unified representation space that bridges 3D poses and texts, enabling seamless interactions across modalities. Additionally, the mixture-of-visual encoder captures intricate pose details, thereby enhancing fine-grained pose perceptions. The mixed-attention mechanism further enhances pose generation quality while significantly accelerating inference speed. Extensive evaluations across various pose-relevant tasks demonstrate the effectiveness of UniPose in pose comprehension, generation, and editing. % Moreover, UniPose exhibits zero-shot generalization abilities, enabling it to tackle previously unseen pose-relevant tasks.

{
    \small
    \normalem
    \bibliographystyle{ieeenat_fullname}
    \bibliography{main}
}

\clearpage
\setcounter{page}{1}
% \maketitlesupplementary
\setcounter{section}{0}
\setcounter{table}{0}
\setcounter{figure}{0}
\twocolumn[{
\renewcommand\twocolumn[1][]{#1}
\maketitlesupplementary
\begin{center}
    \captionsetup{type=figure}
    \includegraphics[width=0.8\textwidth]{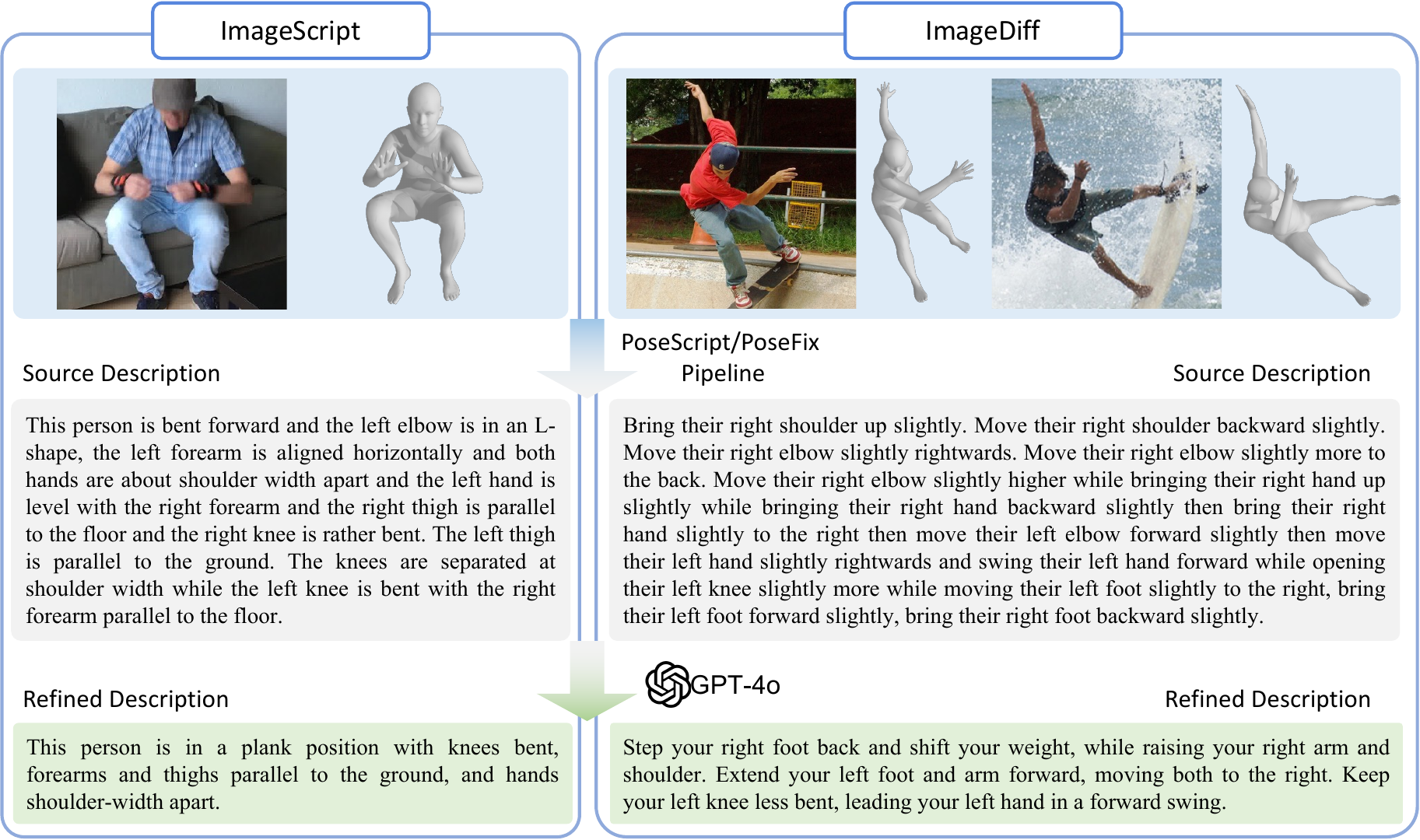}
    \captionof{figure}{The annotation workflow for ImageScript (left) and ImageDiff (right) datasets. }
    \label{fig:data-pipeline}
\end{center}
}]

\begin{figure*}[htbp]
 \centering
 \includegraphics[width=0.8\linewidth]{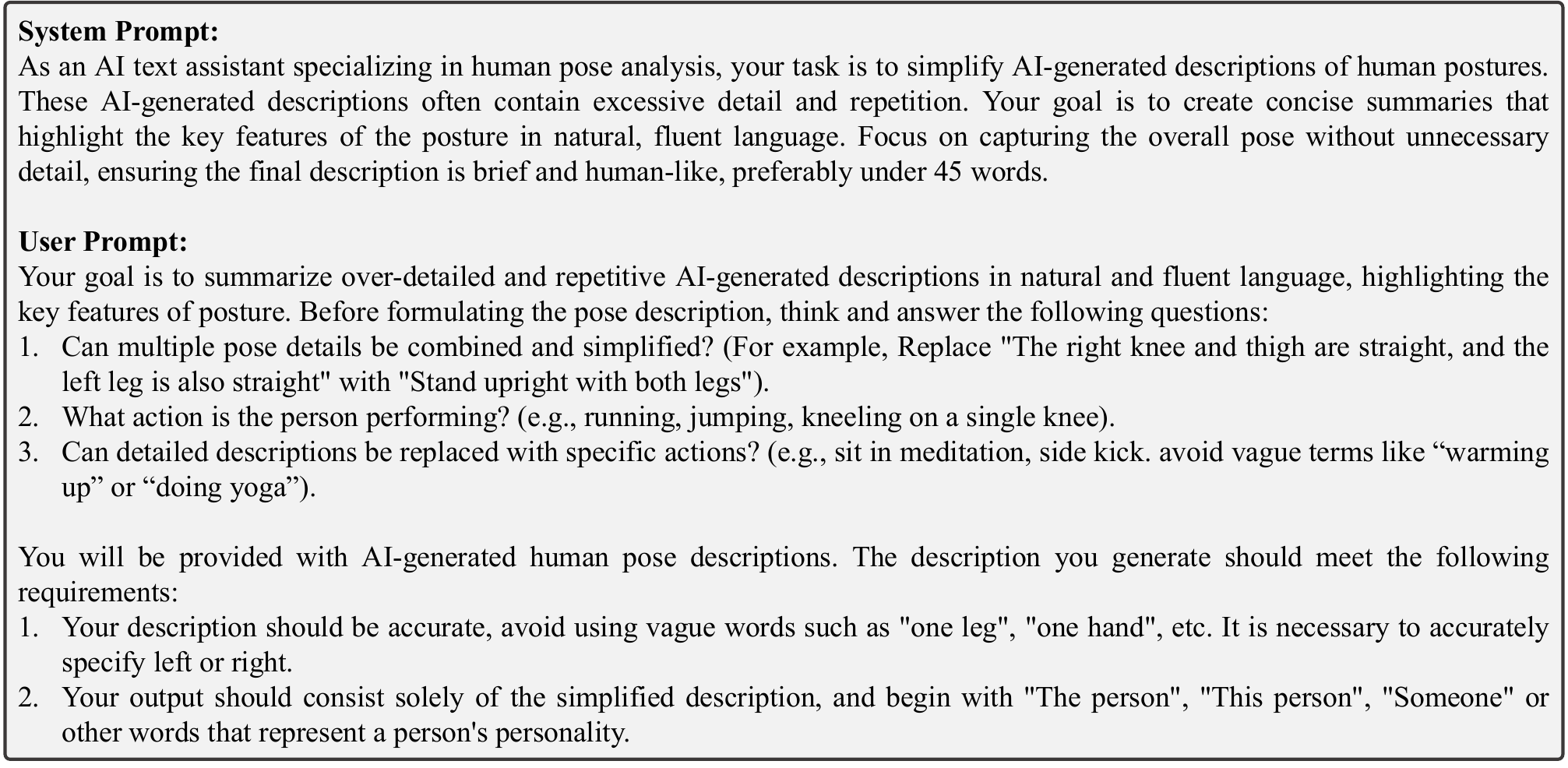
 }
 \caption{Prompt to query GPT-4 for refining text in the ImageScript dataset.}
 \label{fig:image-script-prompt}
\end{figure*}

\begin{figure*}[htbp]
 \centering
 \includegraphics[width=0.8\linewidth]{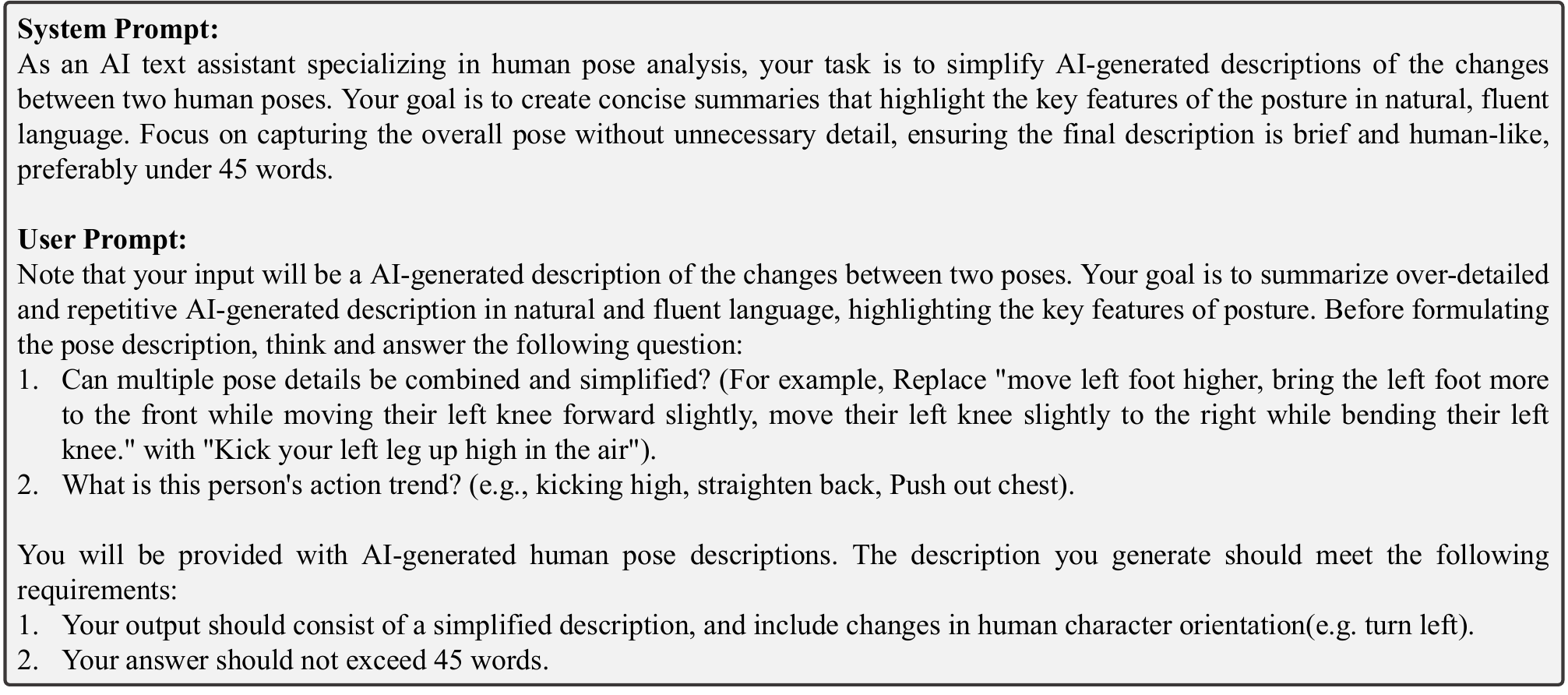}
 \caption{Prompt to query GPT-4 for refining text in the ImageDiff dataset.}
 \label{fig:image-diff-prompt}
\end{figure*}

\begin{table*}
\centering
\footnotesize
\centering
\begin{tabular}{c|ccc}
\toprule
Training paradigm & Task & Dataset & Samples \\
\midrule
\multirow{2}{*}{\makecell{Pose-Text Align \\ Pretraining}} & \multirow{2}{*}{\makecell{Pose-to-Text,  Pose-Diff, \\ Text-to-Pose, Pose-Edit}} & PoseScript-A & 70k \\
 &  & PoseFix-A & 93k \\
\midrule
\multirow{3}{*}{\makecell{Visual Projector \\ Pretraining}} & \multirow{3}{*}{\makecell{Image-to-Text, \\ Image-Diff, \\ Pose Estimation}} & ImageScript-A & 50k \\
 &  & ImageDiff-A & 50k \\
 &  & PoseEst & 100k \\
\midrule
\multirow{5}{*}{\makecell{Instruction \\ Finetuning}} & \multirow{5}{*}{All tasks} & PoseScript-H & 5k \\
 &  & PoseFix-H & 5k \\
 &  & ImageScript-R & 6k \\
 &  & ImageDiff-R & 6k \\
 &  & PoseEst & 6k \\
 \bottomrule
\end{tabular}
\caption{\textbf{Detailed datasets for training UniPose}. The PoseScript dataset provides human annotations (PoseScript-H) and expands its dataset with automated captions (PoseScript-A), as does the PoseFix dataset.}
\label{tab:tasks}
\end{table*}

\begin{table*}[]
\centering
\footnotesize
\begin{tabular}{l|c|c|c}
\toprule
\multicolumn{1}{l|}{Task} & Sub-Task & Input & OutPut \\ \hline
\multicolumn{1}{l|}{\multirow{8}{*}{\makecell{Pose \\ Comp}}} & \multirow{2}{*}{Pose-to-Text} & Generate a description of the SMPL pose: {\ttfamily <pose>}. & \multirow{8}{*}{{\ttfamily <caption>}} \\ 
\multicolumn{1}{l|}{} &  & Interpret the SMPL pose in {\ttfamily <pose>} and generate a written description. &  \\ \cmidrule{2-3}
\multicolumn{1}{l|}{} & \multirow{2}{*}{Pose-Diff} & Provide a summary of how SMPL pose {\ttfamily <pose>} differs from {\ttfamily <pose>}. &  \\ 
\multicolumn{1}{l|}{} &  & Detail any SMPL pose changes seen between {\ttfamily <pose>} and {\ttfamily <pose>}. &  \\ \cmidrule{2-3}
\multicolumn{1}{l|}{} & \multirow{2}{*}{Image-to-Text} & Describe the pose of the individual in the {\ttfamily <image>}. &  \\
\multicolumn{1}{l|}{} &  & Analyze {\ttfamily <image>} and describe the posture displayed. &  \\ \cmidrule{2-3}
\multicolumn{1}{l|}{} & \multirow{2}{*}{Image-Diff} & Compare {\ttfamily <image>} and {\ttfamily <image>}, outline how the person's posture differs. &  \\
\multicolumn{1}{l|}{} &  & Identify how the individual's pose varies from {\ttfamily <image>} to {\ttfamily <image>}. &  \\ \hline
\multicolumn{1}{l|}{\multirow{4}{*}{\makecell{Pose \\ Gen}}} & \multirow{2}{*}{Pose Estimation} & Could you estimate the SMPL pose of the individual in {\ttfamily <image>}? & \multirow{6}{*}{{\ttfamily <pose>}} \\
\multicolumn{1}{l|}{} &  & Look at the {\ttfamily <image>} and return the SMPL pose parameters for the figure shown. &  \\ \cmidrule{2-3}
\multicolumn{1}{l|}{} & \multirow{2}{*}{Text-to-Pose} & Could you generate the SMPL pose from the description: {\ttfamily <caption>}? &  \\
\multicolumn{1}{l|}{} &  & Using the description {\ttfamily <caption>}, please create the corresponding SMPL pose. &  \\ \cmidrule{1-3}
\multicolumn{2}{c|}{\multirow{2}{*}{Pose Editing}} & Modify {\ttfamily <pose>} based on this instruction: {\ttfamily <caption>}. &  \\
\multicolumn{2}{c|}{} & Refine {\ttfamily <pose>} by applying the description provided: {\ttfamily <caption>}. & \\
\bottomrule
\end{tabular}
\caption{Examples of instruction templates utilized during the instruction finetuning stage of UniPose training.}
\label{tab:instruct-finetune}
\end{table*}
\renewcommand{\thesection}{\Alph{section}}
\renewcommand{\thesubsection}{\Alph{section}.\arabic{subsection}}
\vspace{-4mm}
In this Appendix, we present a comprehensive overview of UniPose, covering its datasets, implementation details, performance evaluation, and limitations. First, we introduce two new image-text datasets, ImageScript and ImageDiff, along with a detailed description of the training data used for UniPose (Sec. \ref{data-collection}). Next, we outline the implementation details of the pose tokenizer, retrieval models, and UniPose, including their architectural designs and training configurations (Sec. \ref{impl-details}). Additionally, we present experimental results to evaluate the performance of the tokenizer and retrieval models (Sec. \ref{experiments}). Finally, we offer additional qualitative results (Sec. \ref{qual-eval}) and conclude with an analysis of UniPose’s limitations (Sec. \ref{limiation}).

\section{Data Collection}
\label{data-collection}
To address the lack of datasets combining human images with pose descriptions, we present the ImageScript and ImageDiff datasets, specifically designed to bridge this gap in visual-textual pose comprehension.

\subsection{ImageScript}

ImageScript dataset aims to provide accurate and detailed textual descriptions of human poses depicted in images. Existing pose estimation datasets, collectively referred to as PoseEst (\textit{e.g.}, Human3.6M \cite{ionescu2013human3.6m}, MPI-INF-3DHP \cite{mehta2017mpi-inf-3dhp}, COCO \cite{lin2014coco}, MPII \cite{andriluka20142d-mpii}, and 3DPW \cite{von2018-3dpw}) offer precise human poses paired with images. PoseScript \cite{delmas2024posescript} introduces a pipeline for automatically generating textual descriptions of human poses. Building on these efforts, our ImageScript dataset integrates human images, poses, and detailed textual descriptions to advance visual-textual pose comprehension.

The ImageScript dataset comprises 52k image-text pairs, with the images sourced from the PoseEst datasets. Following PoseScript \cite{delmas2024posescript}, we first normalize the joint positions of each pose annotation from PoseEst datasets using the neutral SMPL body model \cite{smpl}, employing default shape coefficients and a global orientation of 0. To ensure diversity, we apply the farthest point sampling algorithm to select samples using the mean per joint error (MPJE) as the distance metric. Starting with a randomly selected pose, we iteratively add the pose with the highest MPJE to the selected set until the desired sample size is reached. 

For textural annotations, we utilize the automatic pipeline from PoseScript to generate three diverse captions for each sampled pose. However, automatically generated captions often contain excessive detail and repetition, lacking the simplicity and fluency characteristic of human language. To address this, we use GPT-4 \cite{2023arXivgpt4} to refine the captions, transforming verbose and redundant descriptions into concise, natural expressions. Details of the query prompt and the annotation workflow are provided in Fig. \ref{fig:data-pipeline} and Fig. \ref{fig:image-script-prompt}, respectively. 

\noindent
\textbf{Dataset statistics}. The dataset generated using PoseScript's automatic pipeline is referred to  ImageScript-A, while the GPT-4-refined version is named ImageScript-R. Image-pose pairs are initially sampled from Human3.6M (15k), MPI-INF-3DHP (25k), COCO (5k), and MPII (5k) datasets. Textual pose descriptions for each pose are then generated using the automatic pipeline, forming the ImageScript-A dataset. To construct the ImageScript-R training set, 6,250 examples are uniformly sampled from ImageScript-A. Additionally, 2000 samples from the 3DPW dataset are selected to create the ImageScript-R test set. The captions in ImageScript-R are refined using GPT-4, transforming the automatically generated descriptions into more concise and natural expressions.

\subsection{ImageDiff}

ImageDiff dataset is designed to provide textual descriptions of human pose differences between image pairs, enabling the model to effectively perceive and interpret pose variations across different visual inputs. Building on PoseFix \cite{delmas2023posefix}, which introduced a pipeline for automatically generating comparative descriptions for 3D SMPL pose pairs, we propose ImageDiff, a dataset comprising image pairs, corresponding 3D pose pairs, and textual descriptions of pose differences.

The ImageDiff dataset consists of 52k triplets in the form of \textit{\{image A, image B, text\}}, where the text describes how to modify the human pose from image A (the source image) to match image B (the target image). The corresponding pose annotations for images A and B are denoted as poses A and B. The process for selecting image B is consistent with the approach used in the ImageScript dataset. For selecting image A, following PoseFix \cite{delmas2023posefix}, we first calculate the cosine similarity between the pose retrieval features (Sec. \ref{retrieval model}) of each pose B and all other poses in the PoseEst datasets. The top 100 poses with the highest similarity are shortlisted as candidates for pose A. To ensure diversity, we leverage posecode information \cite{delmas2024posescript} to verify that each pose pair exhibits at least 10 distinct low-level pose properties. 

The pose difference descriptions are generated using the automatic annotation pipeline from PoseFix, producing three captions for each sampled pose pair. Similar to ImageScript, we use GPT-4 to refine these captions, transforming the automatically generated annotations into concise, easy-to-read descriptions. The query prompt and annotation workflow are detailed in Fig. \ref{fig:data-pipeline} and Fig. \ref{fig:image-diff-prompt} respectively.

\noindent
\textbf{Dataset statistics}. The dataset generated using PoseFix's automatic pipeline is referred to as ImageDiff-A, while the GPT-4-refined version is named ImageDiff-R. Images B are initially sampled from Human3.6M (15k), MPI-INF-3DHP (25k), COCO (5k), and MPII (5k) datasets, following the same setup as ImageScript-A. Images A are subsequently selected from the corresponding dataset following the method mentioned above. The human pose difference descriptions for each image pair are then generated via the automatic pipeline to construct ImageDiff-A. For ImageDiff-R, 6,250 examples are uniformly sampled from ImageDiff-A to form the training set, and 2000 image pairs are sampled from the 3DPW dataset for the test set. Finally, GPT-4 is employed to refine the text descriptions in ImageDiff-R.

\subsection{Training Data Details}

We employ specific tasks and datasets for each training stage of UniPose, as summarized in Tab. \ref{tab:tasks}. In details:

\begin{itemize}
    \item \textbf{Pose-Text Alignment Pretraining Stage.} We incorporate four pose-text-related tasks: two pose comprehension tasks (Pose-to-Text and Pose-Diff), one pose generation task (Text-to-Pose), and the Pose-Edit task. Drawing inspiration from the success of PoseScript \cite{delmas2024posescript} and PoseFix \cite{delmas2023posefix} in leveraging automatic captioning pipelines to scale datasets, we use PoseScript-A and PoseFix-A, both rich in automatically generated captions, as the training set. This extensive data effectively facilitates the alignment of pose and text modalities.

    \item \textbf{Visual Projector Pretraining Stage.}  We include three image-related tasks: two pose comprehension tasks (Image-to-Text and Image-Diff), and one pose generation task (Image-to-Pose), using ImageScript-A, ImageDiff-A, and the PoseEst datasets for training.

    \item \textbf{Instruction Fine-tuning Stage.} In this stage, the model is trained across all tasks to ensure it understands and generates text aligned with human expression. The training process uses the PoseEst dataset,  human-annotated datasets such as PoseScript-H and PoseFix-H, and GPT-refined datasets like ImageScript-R and ImageDiff-R.  Additionally, we design task-specific instruction templates to enhance UniPose’s instruction-following capabilities, detailed in Tab. \ref{tab:instruct-finetune}.
\end{itemize}

\begin{table}[]
\centering
\footnotesize
\begin{tabular}{c|ccc}
\toprule
Configuration & \makecell{Pose-Text Align \\ Pretraining} & \makecell{Visual Projector \\ Pretraining} & \makecell{Instruction \\ Finetuning} \\
\midrule
Batch Size & 24 & 8 & 8 \\
Learning Rate & 1.5e-4 & 5e-5 & 5e-5 \\
Epochs & 6 & 2 & 2 \\
Image Res & \multicolumn{3}{c}{336 $\times$ 336 / 256 $\times$ 256} \\
Patch Size & \multicolumn{3}{c}{14 $\times$ 14 / 16 $\times$ 16} \\
Warmup Epochs & \multicolumn{3}{c}{0.03} \\
LR Schedule & \multicolumn{3}{c}{Cosine} \\
Optimizer & \multicolumn{3}{c}{AdamW} \\
\bottomrule
\end{tabular}
\caption{
\textbf{Training hyperparameters of UniPose.} Image Res denotes the input image resolution of CLIP-ViT and Pose-ViT, and the same as Patch Size. 
}
\label{tab-unipose-hyper-parameter}
\end{table}

\begin{table*}[]
\centering
\footnotesize
\begin{tabular}{l|ccc|ccc|c}
\toprule
\multirow{2}{*}{Method} & \multicolumn{3}{c|}{$R^{P2T} \uparrow$} & \multicolumn{3}{c|}{$R^{T2P} \uparrow$} & \multirow{2}{*}{mRecall} \\  \cmidrule{2-7}
& Top-1 & Top-5 & Top-10 & Top-1 & Top-5 & Top-10 & \\
\midrule
\multicolumn{8}{l}{Pose-Text Retrieval} \\
\midrule
PoseScript & 22.3 & 50.1 & 62.9 & 22.1 & 51.4 & 63.1 & 45.3 \\
UniPose & \textbf{31.3} & \textbf{60.1} & \textbf{73.0} & \textbf{31.4} & \textbf{62.5} & \textbf{73.8} & \textbf{55.5} \\
\midrule
\multicolumn{8}{l}{Pose Pair-Text Retrieval} \\
\midrule
PoseFix & 13.9 & 33.2 & 45.2 & 14.1 & 30.1 & 42.5 & 30.0 \\
UniPose & \textbf{15.7} & \textbf{34.0} & \textbf{44.7} & \textbf{15.2} & \textbf{34.0} & \textbf{44.6} & \textbf{31.3} \\
\bottomrule
\end{tabular}
\caption{\textbf{The retrieval results on the PoseScript \cite{delmas2024posescript} and PoseFix \cite{delmas2023posefix} datasets.} We report Top 1 / 5 / 10 $R^{P2T}$ and $R^{T2P}$, 
% representing the text-to-pose and pose-to-text retrieval recall, respectively, 
along with the mean recall (mRecall), which is the average of all retrieval recall values.}
\label{tab:retrieval}
\end{table*}

\begin{table}[]
\footnotesize
\centering
\begin{tabular}{l|cc|cc}
\toprule
\multirow{2}{*}{} & \multicolumn{2}{c}{AMASS $\downarrow$} & \multicolumn{2}{c}{MOYO $\downarrow$} \\ \cmidrule{2-5}
 & MPJPE & PA-MPJPE & MPJPE & PA-MPJPE \\
\midrule
w/o. Noise & 6.7 & 3.8 & 32.6 & 11.7 \\
w/. Noise & \textbf{6.2} & \textbf{3.7} & \textbf{23.1} & \textbf{11.3} \\
\bottomrule
\end{tabular}
\caption{\textbf{Ablation on global orientation noise for the Pose Tokenizer.} }
\label{tab:pose-tokenizer}
\end{table}

\section{Implementation details}
\label{impl-details}
\subsection{Pose Tokenizer}

We provide a detailed explanation of the training objectives for the pose tokenizer. The pose tokenizer is trained using reconstruction loss $\mathcal{L}_{r}$, embedding loss $\mathcal{L}_e$, and commitment loss $\mathcal{L}_c$. To further improve the generated pose quality, we utilize vertices and position regularization in the reconstruction loss, as follows:
\begin{equation}
\begin{aligned}
&\mathcal{L}_{vq}  = \mathcal{L}_{r} + \mathcal{L}_e + \mathcal{L}_c,  \ \text{where}, \\
&\mathcal{L}_r = \lambda_{1}\left\| \widehat{\boldsymbol{p}} -\boldsymbol{p}\right\|_2 + \lambda_{2}\left\| \widehat{\boldsymbol{v}} -\boldsymbol{v}\right\|_2 + \lambda_{3}\left\| \widehat{\boldsymbol{j}} -\boldsymbol{j}\right\|_2, \\
&\mathcal{L}_e = \left\|sg\left[\boldsymbol{z}\right] - \widehat{\boldsymbol{z}}\right\|^2_2, \quad \mathcal{L}_c = \left\|\boldsymbol{z} - sg\left[\widehat{\boldsymbol{z}}\right]\right\|^2_2, 
\label{eq2} 
  \end{aligned}
\end{equation} 
where $\boldsymbol{v}$ and $\boldsymbol{j}$ denotes the ground truth SMPL mesh vertices and joints positions derived from $\boldsymbol{p}$, $\widehat{\boldsymbol{v}}$ and $\widehat{\boldsymbol{j}}$ denotes the predicted vertices and positions derived from $\widehat{\boldsymbol{p}}$, $sg[\cdot]$ is the stop gradient operator, and $\lambda_1$, $\lambda_2$ and $\lambda_3$ are the weighting factors. 

\noindent
\textbf{Training Configurations.} 
For the training of Pose Tokenizer, we use AdamW as the optimizer with a batch size of 256 and an initial learning rate of 2e-4. The model is trained for 240 epochs and the weighting factors $\lambda_1$, $\lambda_2$ and $\lambda_3$ are set to $20$, $100$, $100$ respectively. We set the codebook size to 2048, representing each 3D pose with 80 discrete tokens. 
Following TokenHMR \cite{dwivedi2024tokenhmr}, we augment random joints with noise starting at 0.01, progressively increasing after every 5K iterations. To further enhance robustness to global orientation variations, we introduce random perturbations of -45 to 45 degrees in the z-direction and -20 to 20 degrees in the x and y directions. The effect of global orientation noise is analyzed in Sec. \ref{experiments}.

\subsection{Retrieval Model}
\label{retrieval model}

To compute the Pose-Text retrieval metric, a retrieval model is required to rank a large collection of poses based on their relevance to a given textual query, and vice versa. 

\noindent\textbf{Pose-Text Retrieval Model} consists of a pose encoder and a text encoder. For pose feature extraction, we directly employ the pose encoder from the pose tokenizer and add 1D Conv for dimensionality reduction. For the text encoder, we use a bidirectional GRU \cite{cho2014bigru} with one layer for text feature extraction, with word embeddings and the text tokenizer derived from a pretrained DistilBERT \cite{sanh2019distilbert} model. Both pose and text are encoded into 512-dimensional feature vectors. Following PoseScript \cite{delmas2024posescript}, we adopt the Batch-Based Classification (BBC) loss as the training objective:
\begin{equation}
    \mathcal{L}_{BBC} = -\frac{1}{B}\sum_{i=1}^{B}\log\frac{\exp(\gamma(x_i, y_i))}{\sum_{j}\exp(\gamma\delta(x_i, y_j))}
\end{equation}
where $\gamma$ is a learnable temperature parameter, $\delta$ is the cosine similarity function, and $(x_i, y_i)$ denotes pose-text pairs.

\noindent\textbf{Pose Pair-Text Retrieval Model} is designed for retrieving pose pairs and text in the Pose/Image-Diff task. Its architecture is similar to the pose-text retrieval model, with the key difference being that the pose encoder processes each pose in the pair separately. The extracted features are concatenated along the channel dimension and passed through multiple 1D Conv layers for dimensionality reduction. Both the pose encoder and text encoder generate 512-dimensional feature vectors, utilizing the same training objective as the Pose-Text retrieval model. 

\noindent
\textbf{Training Configurations.}
Following PoseScript and PoseFix, the retrieval models are first pretrained on automatically generated captions (PoseScript-A and PoseFix-A) and then fine-tuned on human-written captions (PoseScript-H and PoseFix-H). The retrieval models are trained for 120 epochs across the pretraining and fine-tuning stages. We use the Adam optimizer, with a batch size of 512 for pretraining and 32 for fine-tuning. The learning rate is set to 2e-4, and the learnable temperature parameter $\gamma$ is initialized to 10. In the main text, all experiments use our proposed retrieval model, except for Text-to-Pose task, which utilizes the retrieval model from PoseScript \cite{delmas2024posescript}.

\begin{figure}[htbp]
 \centering
 \includegraphics[width=0.95\linewidth]{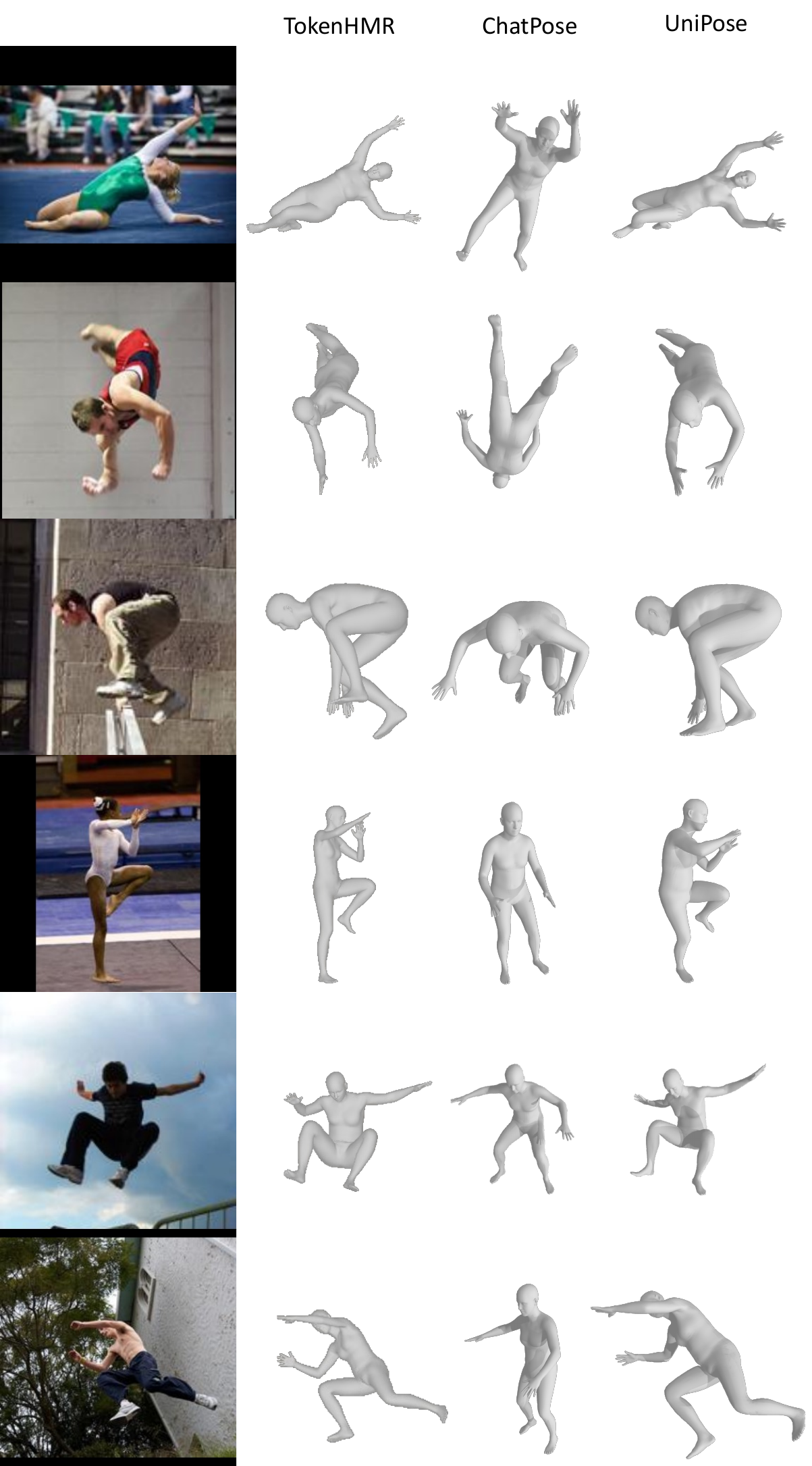}
 \caption{\textbf{Qualitative comparison on pose estimation task}.
 We compare multi-modal LLMs (ChatPose \cite{feng2024chatpose}) and traditional HMR methods (TokenHMR \cite{dwivedi2024tokenhmr}) with our UniPose on LSP \cite{johnson2011lsp-dataset} dataset. }
 \label{fig:pose-est-vis}
\end{figure}

\begin{figure}[htbp]
 \centering
 \includegraphics[width=\linewidth]{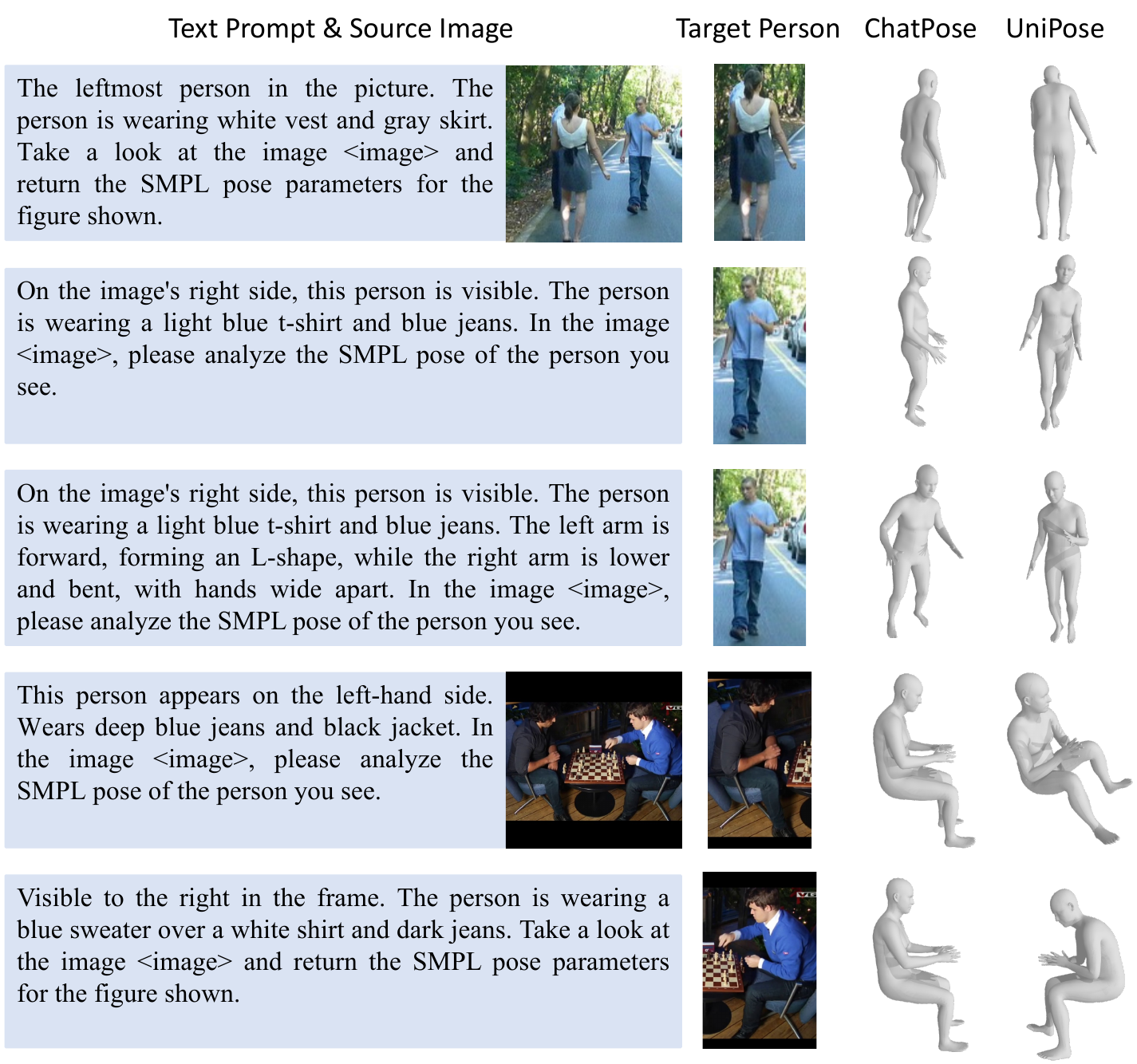}
 \caption{
    \textbf{Qualitative comparison on reasoning-based pose estimation task}. We evaluate the model’s reasoning capabilities in multi-person images.
 }
 \label{fig:additional-finetune}
\end{figure}

\subsection{UniPose}
The detailed training hyperparameter settings for UniPose are provided in Tab. \ref{tab-unipose-hyper-parameter}. In the Pose-Text Alignment Pretraining stage, UniPose is trained for 6 epochs with a batch size of 24 and a learning rate of 1.5e-4. For the Visual Projector Pretraining and Instruction Fine-tuning stages, the model is trained for 2 epochs with a batch size of 8 and a learning rate of 5e-5, respectively. Each stage includes a warm-up period of 0.03 epochs. We adopt the cosine learning rate schedule and use the AdamW optimizer. UniPose incorporates two vision encoders: CLIP-ViT and Pose-ViT, with the input image resolutions and patch sizes of 336 / 14 and 256 / 16 respectively. The output feature map of the Pose-ViT is resized using bilinear interpolation to ensure the visual token count aligns with that of the CLIP-ViT.

\section{Additional Experiments}
\label{experiments}

\subsection{Retrieval Model}

Tab. \ref{tab:retrieval} shows the retrieval results on the PoseScript and PoseFix test sets. All methods are pretrained on automatic captions (PoseScript-A and PoseFix-A) and fine-tuned on human-written captions (PoseScript-H and PoseFix-H). Our Pose-Text retrieval model significantly outperforms PoseScript across all metrics, improving retrieval performance by over 10\%. For Pose Pair-Text retrieval, our model also achieves superior performance. The results demonstrate the effectiveness of our approach in aligning the pose representations with textual descriptions.

\subsection{Pose Tokenizer}

Tab. \ref{tab:pose-tokenizer} illustrates the impact of global orientation noise on the Pose Tokenizer. All methods are trained on the standard training sets of AMASS \cite{mahmood2019amass} and MOYO \cite{tripathi2023moyo}, and evaluated on the AMASS test set and MOYO validation set. The results demonstrate that introducing random noise to global orientation enhances tokenizer robustness, particularly on the MOYO dataset, where MPJPE improves by 9.5. A stronger tokenizer benefits UniPose in handling various pose-related tasks. Therefore, we select the noise-augmented version as the final tokenizer.

\section{Qualitative Evaluation}
\label{qual-eval}

We present the qualitative results of UniPose on pose estimation tasks. In Fig. \ref{fig:pose-est-vis}, we provide visualizations of UniPose’s performance on traditional pose estimation tasks, comparing it with both the traditional method TokenHMR \cite{dwivedi2024tokenhmr} and MLLM-based method ChatPose \cite{feng2024chatpose}. The results show that our approach more accurately estimates human poses, even in scenarios with complex limb articulations.

In Fig. \ref{fig:additional-finetune}, we demonstrate UniPose’s performance on reasoning-based pose estimation tasks. For this, we select 8000 multi-person images from the PoseEst dataset and follow the annotation approach of ChatPose, leveraging GPT-4 \cite{2023arXivgpt4} to label each individual’s behavior, clothes, and pose. Fine-tuning UniPose on this dataset resulted in impressive reasoning capabilities, highlighting the model’s adaptability and generalization to new data.

\section{Limitation}
\label{limiation}

In pose estimation task, the performance of MLLMs-based models still lags behind specialized methods. We argue that these limitations may stem from the constraints imposed by the frozen visual encoder. Future research will focus on developing techniques that enable large language models to more effectively integrate pose-relevant visual features from diverse visual encoders, thereby enhancing their ability to handle complex pose estimation tasks.

% {
%     \small
%     \normalem
%     \bibliographystyle{ieeenat_fullname}
%     \bibliography{main}
% }

\end{document}